\documentclass[sn-apa]{sn-jnl}

\usepackage{graphicx}%
\usepackage{multirow}%
\usepackage{amsmath,amssymb,amsfonts}%
\usepackage{amsthm}%
\usepackage{mathrsfs}%
\usepackage[title]{appendix}%
\usepackage{xcolor}%
\usepackage{textcomp}%
\usepackage{manyfoot}%
\usepackage{booktabs}%
\usepackage{algorithm}%
\usepackage{algorithmicx}%
\usepackage{algpseudocode}%
\usepackage{listings}%

\usepackage{natbib}
\usepackage{cleveref}
\usepackage{euscript}
\usepackage{float}
\usepackage{soul}
\usepackage[varvw]{newtxmath}
\usepackage{bbm}
\usepackage{xspace}

\theoremstyle{thmstyleone}%
\theoremstyle{thmstyletwo}%

\theoremstyle{thmstylethree}%

\renewcommand{\v}[1]{\boldsymbol{#1}}
\newcommand{\bX}{\v{X}}
\newcommand{\bx}{\v{x}}
\newcommand{\bZ}{\v{Z}}
\newcommand{\bz}{\v{z}}
\newcommand{\bY}{\v{Y}}

\newcommand{\bu}{\v{u}}
\newcommand{\bw}{\v{w}}
\renewcommand{\e}{\mathrm{e}}
\newcommand{\Nor}{\EuScript{N}}
\newcommand{\U}{\EuScript{U}}
\newcommand{\Em}{{\mathbb{E}}}
\newcommand{\Pm}{{\mathbb{P}}}
\newcommand{\di}{\mathrm{d}}
\newcommand{\bv}{\v{v}}
\newcommand{\bV}{\v{V}}

\newcommand{\bQ}{\v{Q}}
\newcommand{\bq}{\v{q}}
\newcommand{\point}{\bV}
\newcommand{\spoint}{\ensuremath{\bv}\xspace}

\newcommand{\argmin}{\mathop{\rm argmin}}

\DeclareMathOperator*{\Circ}{\circ}

\makeatletter
\newcounter{HALG@line}
\renewcommand{\theHALG@line}{\thealgorithm.\arabic{ALG@line}}
\makeatother

\begin{document}

\title[A Flow-Based Generative Model for Rare-Event Simulation]{A Flow-Based Generative Model for Rare-Event Simulation}


\author[1]{\fnm{Lachlan} \sur{Gibson}}\email{l.gibson1@uq.edu.au}

\author*[1]{\fnm{Marcus} \sur{Hoerger}}\email{m.hoerger@uq.edu.au}

\author[1]{\fnm{Dirk} \sur{Kroese}}\email{kroese@maths.uq.edu.au}

\affil*[1]{\orgdiv{School of Mathematics \& Physics}, \orgname{The University of Queensland}, \orgaddress{\city{Brisbane}, \postcode{4072}, \state{QLD}, \country{Australia}}}

\abstract{Solving decision problems in complex, stochastic environments is often achieved by estimating the expected outcome of decisions via Monte Carlo sampling. However, sampling may overlook rare, but important events, which can severely impact the decision making process. We present a method in which a Normalizing Flow generative model is trained to simulate samples directly from a conditional distribution given that a rare event occurs. By utilizing Coupling Flows, our model can, in principle, approximate any sampling distribution arbitrarily well. By combining the approximation method with Importance Sampling, highly accurate estimates of complicated integrals and expectations can be obtained. We include several examples to demonstrate how the method can be used for efficient sampling and estimation, even in high-dimensional and rare-event settings. We illustrate that by simulating directly from a rare-event distribution significant insight can be gained into the way rare events happen.}

\keywords{normalizing flows, neural networks, rare events, simulation}

\maketitle

\section{Introduction}
Many decision problems in complex, stochastic environments \citep{Kochenderfer2015decision} are nowadays solved by estimating the expected outcome of decisions via Monte Carlo simulations \citep{kroese2013handbook, kroese2019data, liu2001monte}. The success of Monte Carlo methods is due to their simplicity, flexibility, and scalability. However, in many problem domains, the occurrence of rare but important events --- events that happen with a very small probability, say less than $10^{-4}$ --- severely impairs their efficiency, for the reason that such events do not show up often in a typical simulation run \citep{arief2021certifiable}. On the other hand, failing to consider such rare events can lead to decisions with potentially catastrophic outcomes, e.g., hazardous behaviour of autonomous cars. By using well-known variance reduction techniques such as Importance Sampling~ \citep{kroese2013handbook, BucklewJames2004ItRE} it is possible to sometimes dramatically increase the efficiency of the standard Monte Carlo method. Nevertheless, there are few efficient methods available that give insights into {\em how} a system behaves under a rare event. An important research goal is thus to find methods that simulate a random process {\em conditionally} on the occurrence of a rare event. In this case, the target sampling distribution is the distribution of the original process conditioned on the rare event occurring. More broadly, for both estimation and sampling problems, the challenge is to identify a ``good'' sampling distribution that closely approximates a target distribution. Certain approximation methods, such as the Cross-Entropy method \citep{deBoer2005,rubinstein2016simulation}, train a parametric model by minimizing the cross-entropy between a distribution family and the target distribution. However, typical parametric models are often not flexible enough to efficiently capture all the complexity of many interesting systems. \citet{botev2007generalized} introduced a Generalized Cross-entropy method, which approximates the target distribution in a non-parametric way. However, the quality of the results is determined by various hyper-parameters, which require expertise to tune.

While many standard methods can identify sampling distributions of
sufficient quality to estimate rare-event probabilities and related quantities of interest, extremely good approximations to the
target conditional distribution are required to be able to explore the
behaviour of the simulated system under rare-event conditions. This can
occasionally be achieved by strategically selecting very specific
distribution families using problem-specific information, but this has
been difficult to achieve with conventional simulation
methods. Therefore, it is desirable to have a more general approach
that can closely approximate any target distribution without relying
on problem-specific knowledge. \citet{Gibson2022}
recently introduced a framework for rare-event simulation using neural
networks that aimed to achieve this. In that framework two Multilayer
Perceptrons are trained simultaneously: the first being a generative
model to represent the sampling distribution, and the second to
approximate the probability distribution of the first. While the
framework had some success in one-dimensional problems, the reliance
on a second network and probability density estimation convoluted the training process and made scaling to higher-dimensional problems difficult. \citet{ardizzone2019guided} and \citet{falkner2022conditioning} explore the use of normalizing flows for rare event sampling in the context of image generation and Boltzmann generators. The approach in \citet{ardizzone2019guided} requires a conditioning network to be pre-trained and embedded into the normalizing flows architecture. \citet{falkner2022conditioning} embeds a bias variable into the flow to learn distributions that are biased towards the region of interest. However, such bias variables can be difficult to construct for more complex problems. In contrast, our model is trained end-to-end and without having to introduce a conditioning variable.

We present a new framework, inspired by \citet{Gibson2022}, in which a Normalizing Flows generative model is trained to learn the optimal sampling distribution and used to estimate quantities such as the rare-event probability. The highly expressive nature of some Normalizing Flows architectures, trained using standard deep learning training algorithms, massively expands the range of learnable distributions, while the invertibility of Normalizing Flows allows the exact generative probability density to be computed without the need of a second network. In Section~\ref{sec:theory} we present the background theory and a description of the training algorithm, and in Section~\ref{sec:results} we present several examples of rare-event simulation using this method. The source-code of our method is available at \url{https://github.com/hoergems/rare-event-simulation-normalizing-flows}.



\section{Theory}\label{sec:theory}
The theory of rare-event simulation \citep{JUNEJA2006291, Rubino2009, BucklewJames2004ItRE}, Importance Sampling \citep{Glynn1989, Tokdar2010, Neal2001} and Normalizing Flows \citep{Kobyzev2020, papamakarios2021normalizing, Rezende2015} is well established. In this section we review the relevant background, formulate the foundation of our method, and present the training procedure.

\subsection{Rare-Event Simulation and Importance Sampling}
First let us introduce a mathematical basis for rare-event simulation
and Importance Sampling. Consider a random object (e.g., random
variable, random vector, or stochastic process), $\bX$, taking values
in some space, $\mathcal{X}$, with probability density function,
$p(\bx)$, that represents the result of a simulation
experiment. Running simulations corresponds to sampling $\bX$.
We consider rare events of the form $\{\bX \in \mathcal{A}\}$, where
$\mathcal{A} \subset \mathcal{X}$
is the set of
simulation outcomes $\bx$ that satisfy a predicate ${S(\bx)\geq
  \gamma}$ for some {\em performance function} 
$S:\mathcal{X}\rightarrow\mathbb{R}$ and {\em level parameter} 
$\gamma\in\mathbb{R}$. The probability $c$ of the rare event $\{\bX
\in \mathcal{A}\}$ can thus be written as 
\[
c := \mathbb{P}(\bX \in \mathcal{A}) = \mathbb{P}(S(\bX) \geq \gamma) = \mathbb{E} [\mathbbm{1}_{\mathcal{A}}(\bX)],
\] where $c$ is very small, but not
0. Here,  $\mathbbm{1}_{\mathcal{A}}(\bx)$ is an indicator function that is 1 when
${\bx\in \mathcal{A}}$ and 0 otherwise, and $\mathbb{E}$ represents the expected value.  The expected number of simulations to sample a single rare event is $1/c$, so simulating rare events by sampling $\bX$ directly quickly becomes computationally infeasible the rarer the event is. Therefore, other techniques, such as Importance Sampling become necessary to simulate and analyse rare events.

 In particular,
suppose that $\bX$ is sampled from a probability density function
$p(\bx)$ and that we wish to estimate the quantity
\[
\ell := \Em [H(\bX)] = \int H(\bx) p(\bx) \, \di \bx
\]
via ``crude'' Monte Carlo: Simulate $\bX_1,\ldots,\bX_n$ independently
from 
$p$, and estimate $\ell$ via the sample average $\sum_{i=1}^n
H(\bX_i)/n$.  
The idea of Importance Sampling is to change the probability
distribution  under which the simulation takes place.
 However, computing the expected value
of a function of $\bX$ while using a different sampling density
$q$,  requires a correction factor of the likelihood ratio $p(\bx)/q(\bx)$,
\begin{equation}\label{eq:expected_lambda} \ell:=\Em_p\left[H(\bX)\right]=\Em_q\left[H(\bX)\frac{p(\bX)}{q(\bX)}\right],
\end{equation}
where $\Em_p = \Em$ and $\Em_q$ represent the expected
values under the two probability models. This relationship allows an unbiased estimate of $\ell$ to be
computed by sampling $\bX$ from  $q$ instead of $p$, via
\[
    \widehat{\ell}:=\frac{1}{n}\sum_{k=1}^n H(\bX_k)\frac{p(\bX_k)}{q(\bX_k)},
\]
where $\bX_1,\ldots,\bX_n$ is an independent and identically
distributed (iid) sample from $q$. Note that any choice for $q$ is
allowed, as long as $q(\bx) \neq 0$ when $p(\bx) \neq 0$. The optimal
sampling distribution for estimating $\ell$ in this way is thus the
distribution that minimizes the variance of the estimator
$\widehat{\ell}$. When $H$ is strictly positive (or strictly
negative), choosing the probability density 
\begin{equation}\label{eq:optimal_dist_H}
    q^*(\bx) :=\frac{p(\bx)H(\bx)}{\ell},
\end{equation}
yields in fact a zero-variance estimator, because then $\widehat{\ell}=\ell$. As a
special case, an unbiased estimator of the rare-event probability $c =
\Pm(\bX \in A)$ can be computed as
\begin{equation}\label{eq:c_estimator}
    \widehat{c}:=\frac{1}{n}\sum_{k=1}^n \mathbbm{1}_{\mathcal{A}}(\bX_k)\frac{p(\bX_k)}{q(\bX_k)},
\end{equation}
and the optimal sampling density $q^*$ is  just the original density
$p$ {\em truncated} to the rare-event region $\mathcal{A}$: 
\begin{equation}\label{eq:rare_event_prob_optimal}
    q^*(\bx)= p^{\mathcal{A}}(\bx) := \frac{p(\bx)\mathbbm{1}_{\mathcal{A}}(\bx)}{c}.
\end{equation}
More generally, if we want to estimate the expectation of $H(\bX)$
{\em conditional} on $\bX \in \mathcal{A}$, that is,
\begin{equation}\label{eq:lc}
    \ell^{\mathcal{A}} := \Em_{p^{\mathcal{A}}}[H(\bX)] = \frac{1}{c}{\Em}_q
    \left[H(\bX)\frac{p(\bX)}{q(\bX)} \mathbbm{1}_{\mathcal{A}}(\bX)\right], 
\end{equation}
then
$\ell^{\mathcal{A}}$ can be estimated via
\begin{equation}\label{eq:lc_estimate}
    \widehat{\ell}^{\mathcal{A}}:=\frac{1}{n \, \widehat{c}}\sum_{k=1}^n
    H(\bX_k)\frac{p(\bX_k)}{q(\bX_k)} \mathbbm{1}_{\mathcal{A}}(\bX_k).
\end{equation}
In this case, the optimal Importance Sampling density is
\begin{equation}\label{eq:rare_event_cond_optimal}
    q^*(\bx):=\frac{p(\bx)H(\bx)\mathbbm{1}_{\mathcal{A}}(\bx)}{\ell
      \, c},
\end{equation}

\subsection{Normalizing Flows}
If it is possible to approximate the target density (e.g., the optimal
Importance Sampling density) closely, then we are able to compute
certain quantities of interest (e.g., $c = \Pm(\bX \in \mathcal{A})$ or $\ell = \Em H(\bX))$ with 
low variance. \emph{Normalizing Flows} is a generative model that
can closely approximate a wide variety of probability
densities. Suppose the random object, $\bX$, can be mapped to another
random object, $\bZ$, taking values in some space $\mathcal{Z}$ of the
same dimensionality, and vice versa, via invertible differentiable functions $\v \psi$
and $\v \phi = \v \psi^{-1}$, so that
\[
\bX = \v \psi(\bZ;\bu)\qquad \text{and} \qquad \bZ = \v \phi(\bX;\bu),
\]
where $\bu$ is a vector representing all, if any, parameters. If, under
the target density, $\v \phi$ maps $\bX$ to a random object $\bZ$ that
has a simple \emph{base distribution}, such as a (multivariate) normal
or uniform distribution, then the target distribution can be sampled
indirectly by sampling from the base distribution and mapping the
result using $\v \psi$. The name of such a generative model is derived
from the idea that a sequence of invertible differentiable
transformations can map even a very complex probability distribution to a simple base distribution, thereby forming a `normalizing flow'. The construction of the `flow' relies on the principle that any composition of invertible differentiable functions will also be invertible and differentiable.

In what follows, we assume  that $\cal X$ is a subset of $\mathbb{R}^n$ and
that $\bX$ and $\bZ$ are
``continuous'' random variables; more precisely, that they have
probability densities $p_{\bX}$ and $p_{\bY}$ with respect
to the Lebesgue measure on $\cal X$. 
Since $\v \psi$ is differentiable, the probability density function
$\bX$ can be expressed in terms of the probability
density function $\bZ$ via the relation 
\begin{equation}
p_{\bX}(\bx) =p_{\bZ}(\v \phi(\bx;\bu))\left\vert \det \mathrm{D}\v
\phi(\bx;\bu)\right\vert,\label{eq:transform1}
\end{equation}
or, in terms of $\bz$:
\begin{equation}
p_{\bX}(\v \psi(\bz;\bu))=p_{\bZ}(\bz)\left\vert \det \mathrm{D}\v \psi(\bz;\bu)\right\vert^{-1}.\label{eq:transform2}
\end{equation}
Here, $\mathrm{D}\v \phi(\bx;\bu)$ is Jacobian matrix of $\v
\phi$ and the absolute value of its determinant is the {\em Jacobian}
of $\v \phi$, and similar for $\v\psi$. If $\v \psi$ is a composition of other invertible differentiable functions,
\[
    \v \psi(\,\cdot\, ;\bu)=\Circ_{i} \v \psi_i(\,\cdot\, ;\bu_i),
\]
then we can use the chain rule to combine the Jacobians as a product,
\[
    \left\vert \det \mathrm{D}\v \psi(\,\cdot\, ;\bu)\right\vert = \prod_i \left\vert\det\mathrm{D}\v \psi_i(\,\cdot\, ;\bu_i)\right\vert.
\]
Therefore, analogous to a feed-forward neural network, the full
Jacobian  can be computed during a single pass as $\v
\psi(\bz;\bu)$ is computed. A similar result holds for the inverse:
$\v \phi$. Functional composition increases the complexity of the
model while  maintaining the accessibility of the Jacobian. In this
way, Normalizing Flows provides the opportunity for highly expressible
generative models, with computationally tractable density
functions. As will be discussed in
Section~\ref{sec:objective}, training the model to learn a target
density from a given function, rather than via training data, requires
the ability to compute $p_{\bX}$ as a differentiable function. Therefore, using a Normalizing Flows model improves and simplifies the framework introduced by \citet{Gibson2022} by eliminating the need to train a second neural network to approximate the density function via kernel density estimation.

\subsection{Coupling Flows}
\begin{figure}[H]
    \centering
    \includegraphics[width=\textwidth]{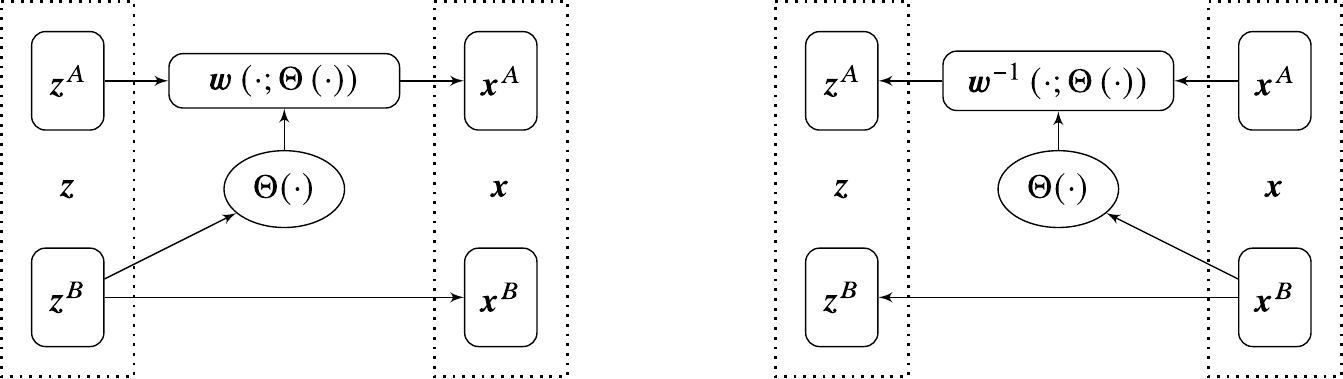}
    \caption{Flow charts showing the structure of a Coupling Flow (left) and its inverse (right).}
    \label{fig:NF_coupling_flow}
\end{figure}

\emph{Coupling Flows}, first introduced by \citet{Dinh2015}, is a family of Normalizing Flows which have been shown to be universal approximators of arbitrary
invertible differentiable functions when
composed appropriately \citep{Takeshi2020}. A single
Coupling Flows unit splits $\bz$ into two vectors, $\bz^A$ and $\bz^B$, as well as $\bx$ into
$\bx^A$ and $\bx^B$. It then transforms one part via an
invertible differentiable function, $\bw$, called the {\em coupling
function}, which contains parameters determined by the other
partition. The relationship between the second part and the
coupling function parameters, called the {\em conditioner}, $\Theta$,
does not need to be invertible, since this part is not transformed and
is accessible when computing the inverse. This means functions with
many learnable parameters, like neural networks, can be used to
capture interesting and complex dependencies between variables. Figure
\ref{fig:NF_coupling_flow} illustrates the structure of a Coupling
Flow models and its inverse.

The Coupling Flow can be expressed mathematically as:

\begin{align*}
    \bx = (\bx^A,\bx^B) = \v \psi(\bz^A,\bz^B ; \bu) &= (\bw(\bz^A      ; \Theta(\bz^B ; \bu)),\bz^B),\\
    \bz = (\bz^A,\bz^B) =\v \phi(\bx^A,\bx^B ; \bu) &= (\bw^{-1}(\bz^A ; \Theta(\bx^B ; \bu)),\bx^B).
\end{align*}
Note that the parameters $\bu$ of the flow are contained in the
conditioner, $\Theta$. Since the $B$-partition does not transform, the
Jacobian of the full transformation is just the Jacobian of the
coupling function; that is, 
\[
\left\vert\det \mathrm{D}\v \psi(\bz,\bu)\right\vert = \left\vert\det \mathrm{D}\bw(\bz^A;\Theta(\bz^B ; \bu))\right\vert.
\]
The coupling
function is often chosen to be a simple affine transformation
\citep{Dinh2015, Dinh2016, NEURIPS2018_d139db6a}. Instead, we base our 
coupling function on the rational function proposed by \citet{ziegler2019}:
\begin{equation}\label{eq:nonlinsquare}
r(z;\theta_1,\theta_2,\theta_3,\theta_4,\theta_5)=\theta_1z+\theta_2+\frac{\theta_3}{1+(\theta_4z+\theta_5)^2}.
\end{equation}
This function has five parameters when applied elementwise. However,
separate parameters can be used for each dimension in the $\bz^A$
partition. For additional complexity, the coupling function can also
be a composition of these rational functions, $w(\cdot;\Theta) =
\Circ_i
r_i(\cdot;\theta_1^i,\theta_2^i,\theta_3^i,\theta_4^i,\theta_5^i)$. In
this case the total number of parameters in the coupling function is
five multiplied by the number of compositions multiplied by the number
of dimension in $\bz^A$. We choose the conditioner function to be a
Multilayer Perceptron, with an input dimensionality to match $\bz^B$ and an output dimensionality to match the number of parameters in the coupling function.

In order for \cref{eq:nonlinsquare} to define an invertible function,
some restrictions have to be placed on the parameters. Choosing
\[
    \theta_1>0,\quad \theta_4>0, \quad \text{and} \quad \vert\theta_3\vert<\frac{8\sqrt{3}\theta_1}{9\theta_4}
\]
ensures that the function $r$ in \cref{eq:nonlinsquare} is a strictly increasing invertible function. We enforce these constraints, following Ziegler and Rush, by processing the output of the conditioner Multilayer Perceptron,
\begin{align*}
    \theta_1&=\e^{\theta_1'}, & \theta_2 &=\theta_2', & \theta_4 &= \e^{\theta_4'}, & \theta_5 &= \theta_5', & \theta_3 &= 0.95\frac{8\sqrt{3}\theta_2}{9\theta_4}\tanh \theta_3',
\end{align*}
where primed letters denote unconstrained outputs of the conditioner
function. The factor of 0.95 in the expression for $\theta_3$ helps
improve stability by precluding barely invertible functions that can
exist near the bound $\vert\theta_3\vert= 8\sqrt{3}\theta_1/(9\theta_4)$. While we do not need to compute the inverse in our method, it can be calculated as the real solution to a cubic equation. 

A single coupling flow unit does not transform the $B$-partition. So, functional composition is necessary to obtain a general approximation. To achieve this, the dimensions of each coupling flow unit are split differently to ensure that all
dimensions have to opportunity to `influence' all other dimensions,
thereby capturing any dependencies between all dimensions. In
practice, we permute the dimensions between Coupling Flow units,  while fixing the partitioned dimensions. The permutation of dimensions can be viewed as special case of a linear transformation, where the transformation matrix is a permutation of the rows of the identity matrix. The Jacobian matrix of a linear transformation is equal to the transformation matrix. So, the Jacobian is just 1 and is trivial to include in the probability density calculations of \cref{eq:transform1} and \cref{eq:transform2}.

\subsection{Objective and Training}\label{sec:objective}
Choosing an appropriate Normalizing Flows model architecture allows
any density to be approximated arbitrarily well if the
correct parameters can be identified. We use a form of stochastic
gradient descent to iteratively tune the parameters of the model to
minimize the Kullback--Leibler (KL) divergence between the model
density, $q$ say, and a known target function, $h:\mathcal{X}\rightarrow
\mathbb{R}$, that is proportional to the target probability density
function. That is, we minimize 
\begin{equation}\label{eq:KL_target}
    \mathcal{D}(q,h) := \Em_q \ln \frac{q(\bX)}{h(\bX)}.
\end{equation}
Note that a model density $q$ that minimizes \cref{eq:KL_target} also
minimizes the KL divergence between $q$ and the actual (normalized) 
target density. The advantage of using \cref{eq:KL_target} is that
the normalization constant of the target density does not need to be known.
This motivates the following minimization program:\footnote{Actually, the first term in this objective corresponds to the negative differential entropy of the base density and does not depend on any parameters, so could be omitted. However, we keep it to make interpreting the loss a little simpler.}
\begin{equation}\label{eq:objective}
    \min_{\bu} L(\bu):=\min_{\bu}\Em\left[\ln p_{\bZ}(\bZ)-\ln \left\vert \det \mathrm{D}\v \psi(\bZ;\bu)\right\vert-\ln h(\v \psi(\bZ;\bu))\right],
\end{equation}
which corresponds to choosing parameters to minimize the KL divergence
between the Normalizing Flows generative density and a target density
whose probability density function is proportional to $h$. The
expression can be derived by substituting $q$ from
\cref{eq:transform2} into \cref{eq:KL_target} and taking
the expectation with respect to the base density. The minimum KL
divergence is zero, when the two densities are identical, so the
minimum value of this objective is the negative natural logarithm of
the normalization constant of $h$. For example, if
$h(\bx)=p(\bx)H(\bx)$, then the minimum value of the objective
function $L$ is $-\ln \ell$. The objective function values can be estimated via
\begin{equation}\label{eq:objective_est}
    \widehat{L}(\bu) := \frac{1}{n}\sum_{k=1}^n \left[\ln p_{\bZ}(\bZ_k)-\ln \left\vert \det \mathrm{D}\v \psi(\bZ_k;\bu)\right\vert-\ln h(\v \psi(\bZ_k;\bu))\right],
\end{equation}
where $\bZ_1,\ldots,\bZ_n$ is an iid sample of the base density.

The objective function $L$ marks a clear distinction between how we
train a Normalizing Flows generative model using a target function,
and how they are typically trained using a data set. In other works,
such as by \citet{Dinh2015}, Normalizing Flows are
trained to learn the density of a data set by maximizing the
log-likelihood of the sampled data. While this can be considered
equivalent to minimizing the KL divergence, the training process
  involves sampling the data set, which corresponds to sampling
  $\bX$, and then moving in the `normalizing' direction with $\v \phi$
  to solve the maximization program:
\[
    \max_{\bu} \Em_q\ln q(\bX)= \max_{\bu} \Em_q\left[\ln p_{\bZ}(\v \phi(\bX;\bu)) + \ln\left\vert \det \mathrm{D}\v \phi(\bX;\bu)\right\vert\right].
\]
After the model is trained, new data can be generated by sampling
$\bZ$ and moving in the `generating' direction using $\bX=\v
\psi(\bZ;\bu)$. Therefore, in the data-focused case, it is essential to be able to compute both transformations, $\v \psi$ and $\v \phi$. However, in the context of rare-event simulation, we do not have access to training data, but a target function that is proportional to a target probability density, and all training occurs in the `generating' direction. Therefore, computing $\v \phi$ is not required (although $\v \psi$ must still be invertible in principle). Algorithm \ref{alg:training} outlines our training procedure.

\begin{algorithm}[H]
    \caption{Training a flow-based generative model using a target function.}\label{alg:training}
    \begin{algorithmic}[1]
    \State Randomly initialize parameters, $\bu$.    
    \For{many iterations}
      \State Independently sample $\bZ_1,\ldots,\bZ_n$ from the base distribution.
      \State Compute $\ln \left\vert \det \mathrm{D}\v \psi(\bZ_k;\bu)\right\vert$ and $\bX_k=\v \psi(\bZ_k;\bu)$ during a single pass.
      \State Compute the target function values $h(\bX_k)$.
      \State Estimate the objective function via \cref{eq:objective_est}.
      \State Adjust parameters $\bu$ via standard gradient descent methods.
    \EndFor
    \end{algorithmic}
\end{algorithm}

\subsection{Conditional Target Densities}
\begin{figure}[H]
    \centering
    \includegraphics[scale = 0.7]{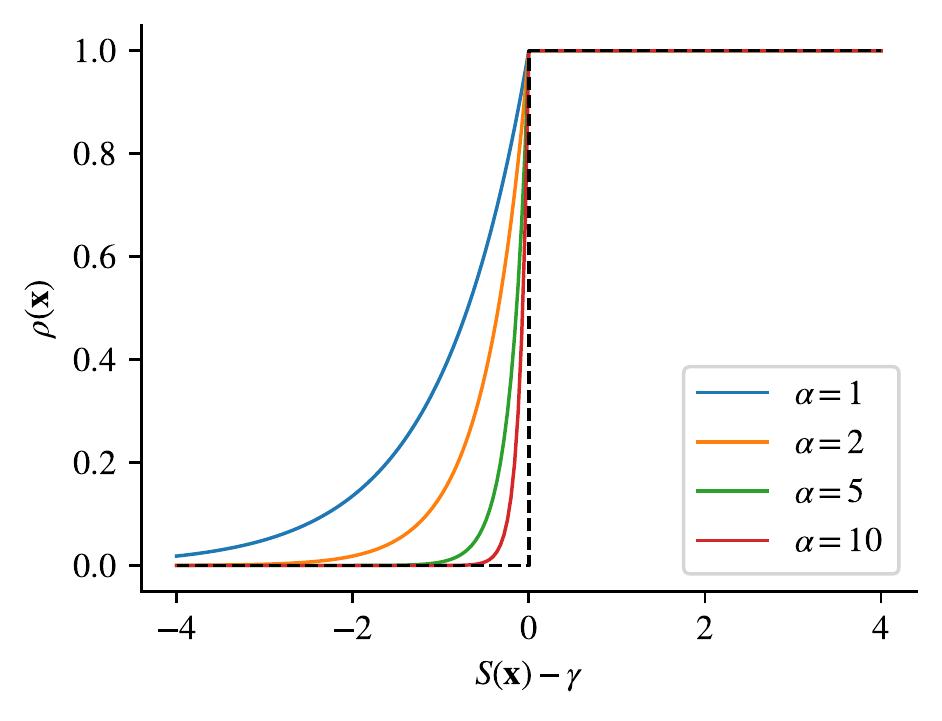}
    \caption{A comparison between the penalty factor $\rho(\bx)$ (colored lines) and the step function $\mathbbm{1}_{\mathcal{A}}(\bx)$ (black dashed line). As $\alpha$ grows the approximation becomes more accurate.}
    \label{fig:penalty_factor}
\end{figure}

As previously shown in \cref{eq:rare_event_prob_optimal}
and \cref{eq:rare_event_cond_optimal}, when estimating rare-event
probabilities and other quantity conditioned on a rare event, the
target density includes an indicator function,
$\mathbbm{1}_{\mathcal{A}}(\bx)$. However, the term, $-\ln h(\v
\psi(\bZ;\bu))$ in the objective function $L$ in  \cref{eq:objective}
requires each value $h(\bx)$ of the target function value to be
strictly positive. To circumvent this problem,  we reuse the solution presented by \citet{Gibson2022}, to approximate $\mathbbm{1}_{\mathcal{A}}(\bx)$ by defining a strictly positive \emph{penalty factor},
\begin{equation}\label{eq:penalty_factor}
    \rho(\bx) := \exp \left[-\alpha\left(\gamma - S(\bx)\right)\mathbbm{1}_{A^c}(\bx)\right],
\end{equation}
which is a positive approximation of $\mathbbm{1}_{\mathcal{A}}(\bx)$ when $\alpha>0$ and $\mathcal{A}^{c}= \{\bx\in\mathcal{X} ~\vert~ S(\bx)< \gamma\}$ is the complement set of $\mathcal{A}$. The approximation is more accurate the larger the value of $\alpha$ and is exact in the limit $\alpha\rightarrow\infty$. Figure \ref{fig:penalty_factor} illustrates the convergence of the approximation.

Therefore, if the target density includes the indicator function as a factor, we replace it with the penalty factor with appropriate performance function, $S$. For example, if the aim is to estimate $\ell^{\mathcal{A}}$ from \cref{eq:lc}, then the target function would be of the form
\[
    h(\bx) = p(\bx)H(\bx) \rho(\bx).
\]
Note that in this case
\[
    \Em\left[-\ln h(\bX)\right]=\Em\left[-\ln p(\bX) H(\bX)\right]-\alpha \Em\left[\left(\gamma - S(\bX)\right)\mathbbm{1}_{A^c}(\bX)\right],
\]
so learning a conditional density by including the penalty factor in the target function is equivalent to learning the unconditional target $p(\bx)H(\bx)$ and including an additional penalty term in the objective that penalizes samples outside the rare-event region. In this way, $\alpha$ can be interpreted as the weight of the penalty. For our experiments we choose $\alpha=100$.

\section{Results}\label{sec:results}
In this section we look at how well Normalizing Flows models can learn
various target functions using several practical examples.

\subsection{Truncated Normal Density}
\begin{figure}[H]
    \centering
    \includegraphics[width = 0.24\textwidth]{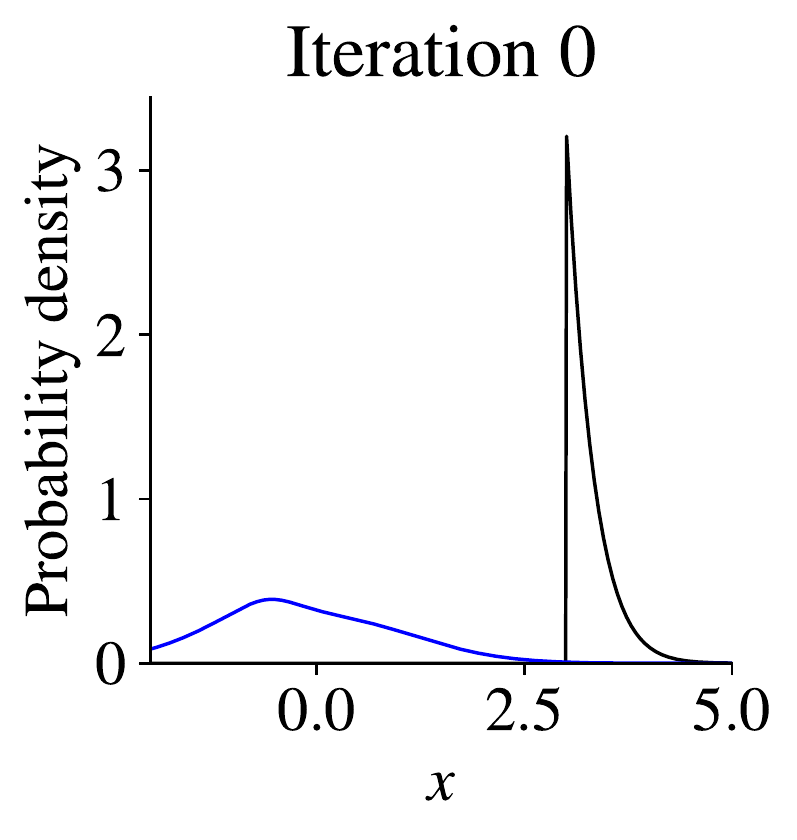}
    \includegraphics[width = 0.24\textwidth]{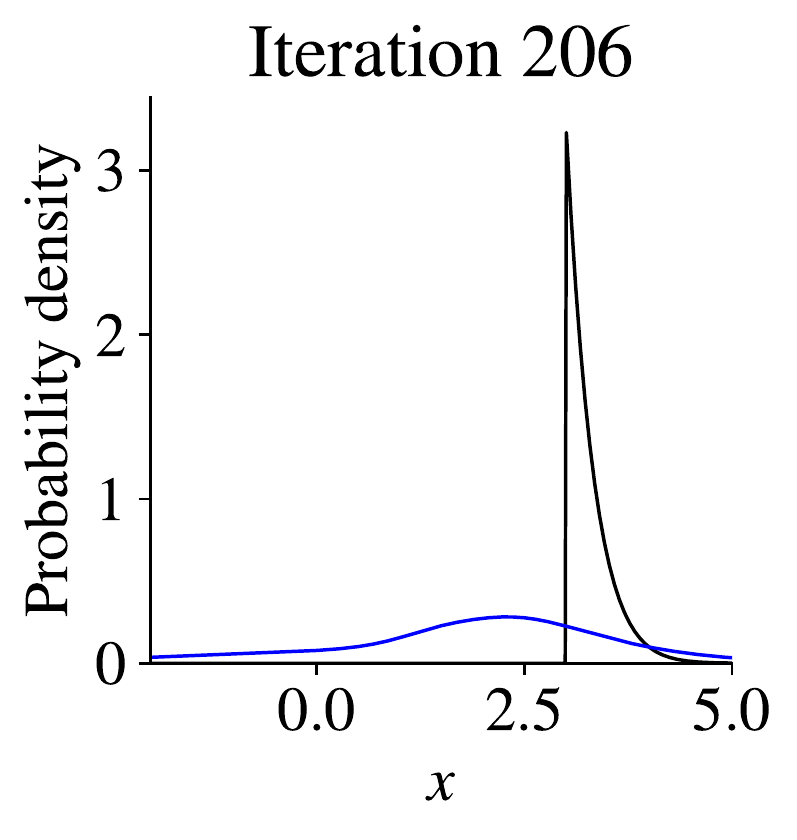}
    \includegraphics[width = 0.24\textwidth]{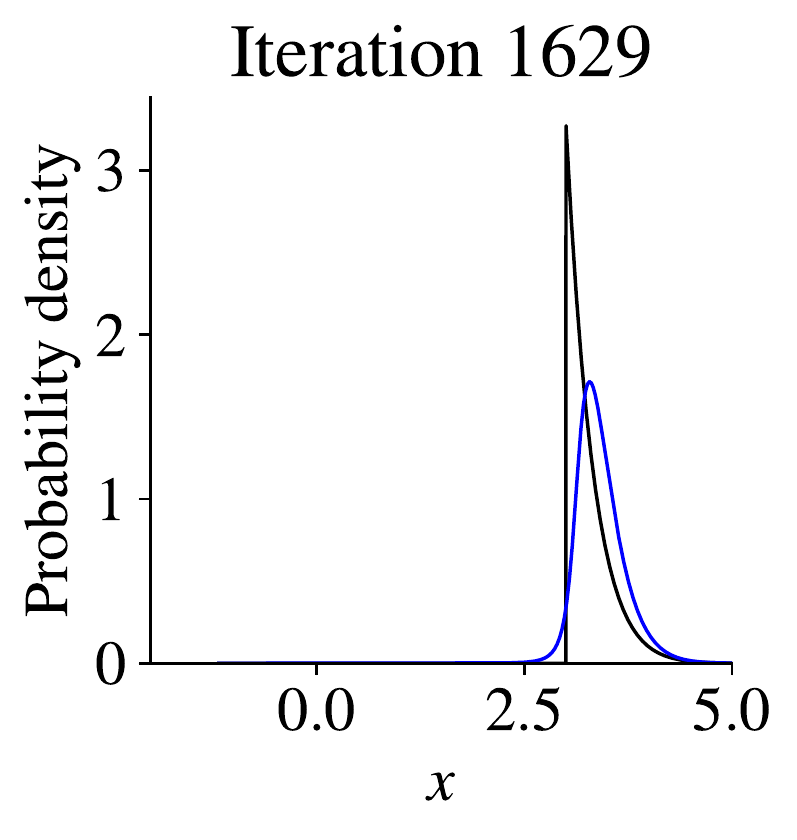}
    \includegraphics[width = 0.24\textwidth]{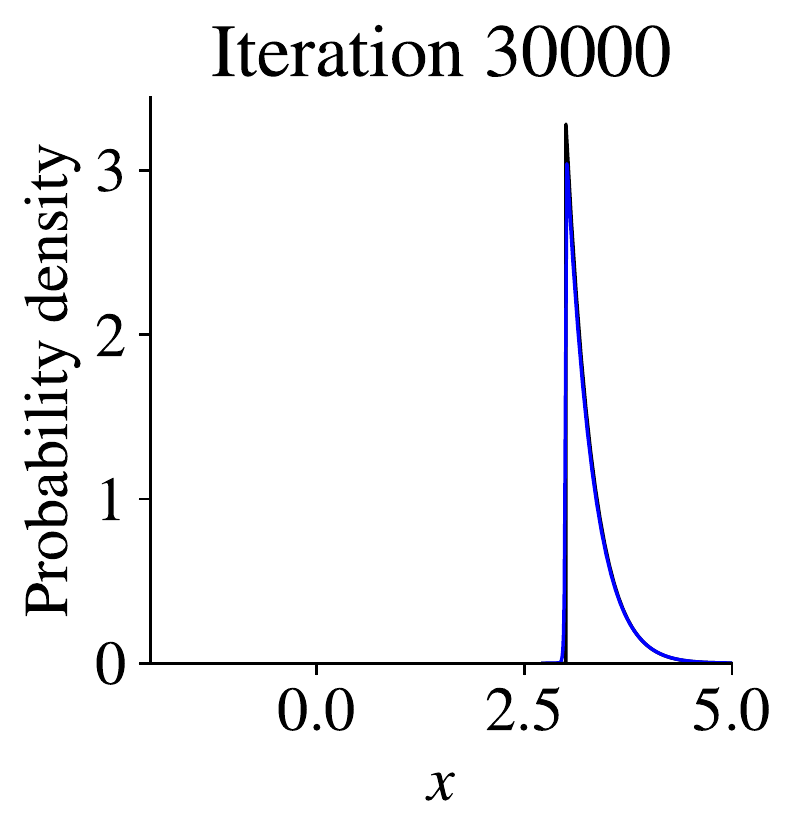}
    \caption{The normalizing flow probability density function at
      different stages of training. The black density is the target
      probability density of a normal density truncated to the
      interval $[3,\infty)$. The blue density is the probability density of $X$ at the corresponding training iteration.}
    \label{fig:trunc_norm}
\end{figure}

Firstly we consider a very simple one-dimensional example of a
truncated normal density. Let $X$ have a standard normal distribution:
$X \sim \EuScript{N}(0,1)$ and consider the rare-event region
$\mathcal{A} = [\gamma, \infty)$. An obvious choice for the performance
  function is $S(x)=x$, so if the goal is to estimate the rare-event
  probability $c = \Pm(X \geq \gamma)$, then the target function is
\[
    h(x) := p_X(x) \exp \left[-\alpha\left(\gamma -
      x\right)\mathbbm{1}_{(-\infty,\gamma)}(x)\right].
\]
A Coupling Flow is unsuitable for a one-dimensional problem, so we use
a composition of three rational functions from
\cref{eq:nonlinsquare}, containing a total of 15 parameters. Choosing
a batch size of $n=1,000$, a learning rate of $0.001$, a weight decay
parameter of $0.0001$ and $\alpha = 100$, the model was trained via
Algorithm~\ref{alg:training} and the Adam gradient descent optimizer
\citep{kingma2017adam} for 30,000 iterations. Figure
\ref{fig:trunc_norm} illustrates how the model converges towards the
truncated normal target density, with threshold parameter $\gamma  =3$.

Using a sample of 1,000 points, \cref{eq:c_estimator} gives
us an estimate 0.00134576 of the rare-event probability $\Pm(X \geq 3)$ , which is about 0.3\% error from the true value of about 0.0013499. From the same sample, the sample standard deviation is about 0.000261. We can estimate the minimum sample size needed for a 1\% standard error by
\[
    n = \left(\frac{\sigma}{\widehat{c}}\frac{1}{0.01}\right)^2\approx 376,
\]
where $\sigma$ is the sample standard deviation of the summand in \cref{eq:c_estimator}. Using a crude Monte Carlo estimator, the relative standard error should be $\sqrt{c(1-c)}\approx 0.0367$, requiring a much larger sample size of about $n=7.4\times 10^6$ for a 1\% standard error.

The learned density is not exactly identical to the target conditional
density, but is very close. To quantify how good the approximation
is, we can estimate the KL divergence between the target density
$q^*$ and the learned density using
\[
    \mathcal{D}(q^*, q)= \Em_{q^*} \left[ \ln \frac{q^*(\bX)}{q(\bX)}
      \right] = \Em_q \left[\frac{q^*(\bX)}{q(\bX)} \ln \frac{q^*(\bX)}{q(\bX)} \right].\label{eq:KL_div_IS}
\]
In this example the KL divergence is estimated to be $0.03465$ with a standard error of $0.56\%$ using sample size 10,000.

\subsection{Two-Dimensional Exponential Density}
\begin{figure}[H]
    \centering
    \includegraphics[width=\textwidth]{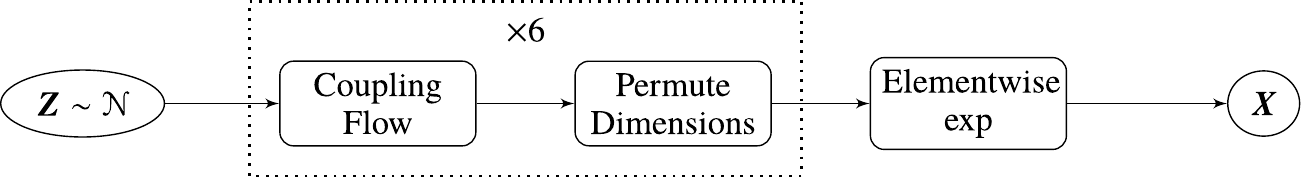}
    \caption{Flow chart showing the normalizing flow structure used in the exponential density example. The base density is a 2D multivariate normal density which forms the input to a sequence of six consecutive Coupling Flow and dimension permutation pairs, followed by an exponential function applied to both dimensions. The learnable parameters are found in the conditioner functions of all six Coupling Flows.}
    \label{fig:flow_chart_sum_exp}
\end{figure}

Secondly, we try another example from \citet{Gibson2022} of a two-dimensional exponential density. In
particular, in this example the probability density function of $\bX =
(X_1,X_2)$ is:
\[
    p(x_1,x_2) = \e^{-(x_1+x_2)}, 
\]
and the rare-event region  is
$\mathcal{A}=\{(x_1,x_2)\in\mathbb{R}^2_+ ~:~ x_1+x_2 \geq \gamma\}$,
with performance function, $S(x_1,x_2)=x_1+x_2$. To estimate $c = \Pm(X_1
+ X_2 \geq \gamma)$, the target function is
\[
    h(x_1,x_2) = \exp \left[-(x_1+x_2)-\alpha\left(\gamma - x_1-x_2\right)\mathbbm{1}_{\{x_1+x_2<\gamma\}}\right].
\]
This time we choose a flow model composed of six Coupling Flows,
interlaced with dimension permutations, followed by an elementwise
exponential function. The exponentiation ensures that the generated
points are positive. The coupling function in each Coupling Flow is a
composition of two rational functions of the form in
\cref{eq:nonlinsquare}. The conditioner in each Coupling Flow is a
linear transformation plus a constant vector. Therefore, the total
number of learnable parameters is 60. Figure
\ref{fig:flow_chart_sum_exp} shows the structure of this model via a flow chart.

\begin{figure}[H]
    \centering
    \includegraphics[scale=0.7]{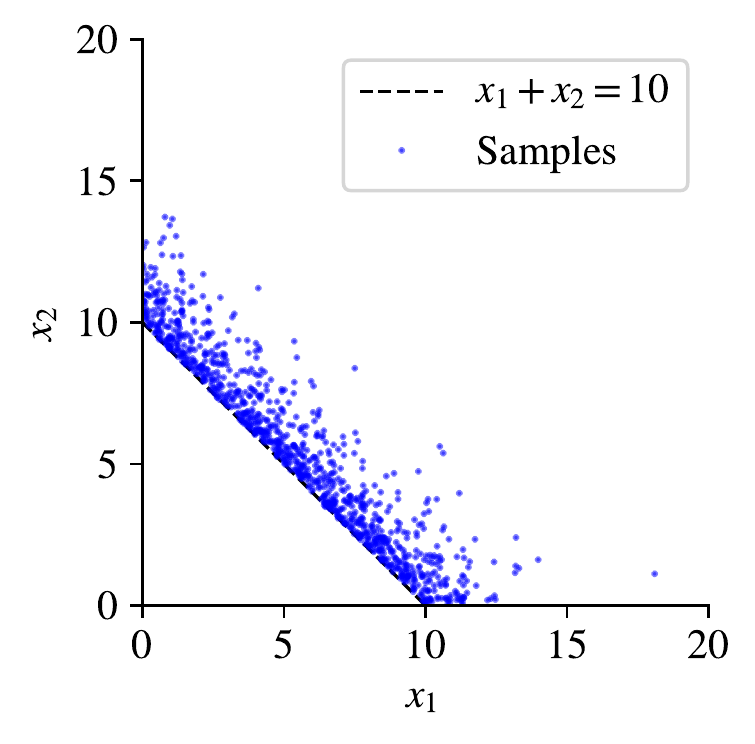}
    \includegraphics[scale=0.7]{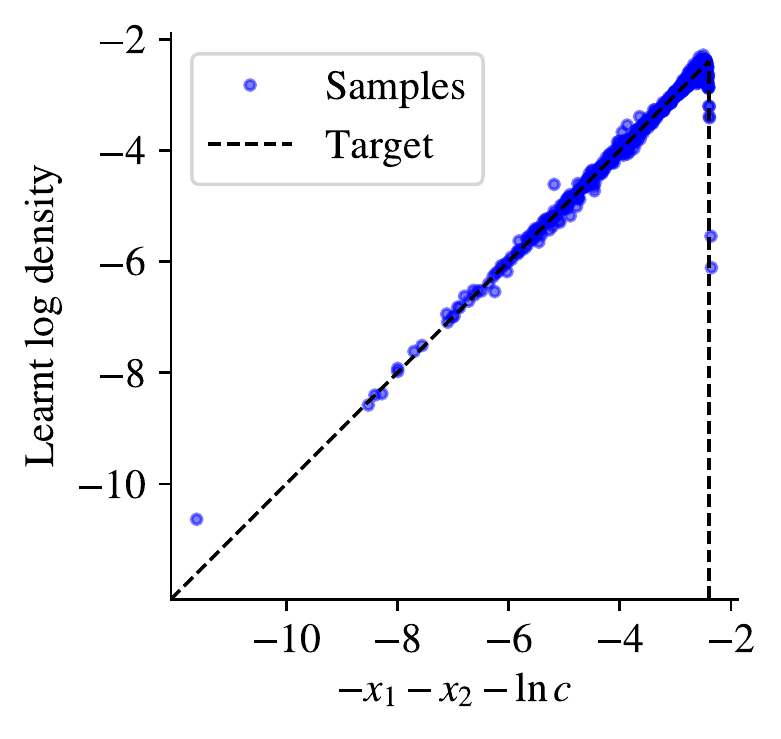}
    \caption{1,000 points sampled from the trained normalizing flow. The left scatter plot shows the how the sampled two-dimensional points almost exclusively satisfy the rare-event condition, $x_1+x_2 \geq 10$. The right plot compares the learned log-density with the renormalized exponential log-density. The dashed black line shows the target density, which is equal to the renormalized log density, until about $-2.398$, which is the maximum log-density when $x_1+x_2 = 10$.}
    \label{fig:sum_exp_scatter}
\end{figure}

The model was trained for 100,000 iterations with a learning rate of
0.0001, a weight decay of constant of 0.0001, and a batch size of 10,000. When $\gamma=10$, $c = (1+\gamma)\e^{-\gamma} \approx 0.000499399$. After training, with a sample size of 1,000 the estimated constant is $0.00049920$, with standard deviation of about $6.34\times 10^{-5}$ giving a relative standard error of 0.40\%. Achieving a comparable standard error using a crude Mote Carlo estimator would require a sample size of more than $10^8$. Figure~\ref{fig:sum_exp_scatter} shows how well the model learned the target density by including a scatter plot of a sample of points as well as comparing the learned probability density with the target probability density. The KL-divergence is estimated to be about 0.011 with 9.7\% standard error using sample size of 10,000.

\subsection{Bridge Network}
\begin{figure}[H]
    \centering
    \includegraphics[width=0.5\textwidth]{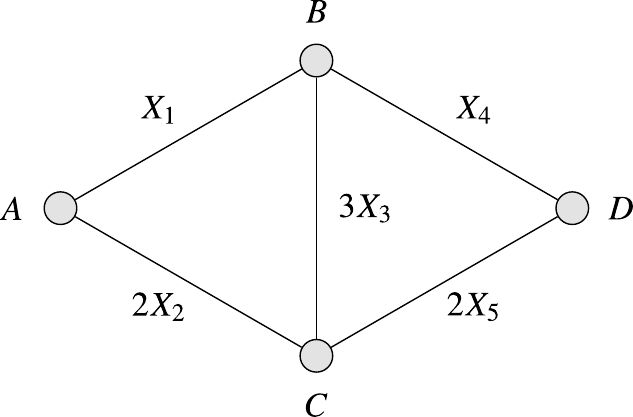}
    \caption{A bridge network containing four nodes and five edges,
      where the edge lengths $X_1,\ldots,X_5$ are random variables.  So the minimum path length from node $A$ to node $D$ is also a random variable.}
    \label{fig:bridge_network}
\end{figure}

This third example comes from Chapter~9 of
\citet{kroese2013handbook}. In this example, we increase the number of
dimensions to five, each corresponding to a random edge length in a
bridge network, shown in Figure~\ref{fig:bridge_network}. The graph contains
five edges and four nodes, where the length of each edge is uniformly
distributed and the goal is to identify the expected shortest path
from node $A$ to node $D$. Specifically, we wish to estimate the
expected length of the shortest path from $A$ to $D$; that is, the
expectation $\ell$ in \cref{eq:expected_lambda}, with $X_i\sim \U(0,1)$ and
\[
    H(\bX) = \min \{X_1+X_4,X_1+3X_3+2X_5,2X_2+3X_3+X_4,2X_2+2X_5\}.
\]
In this case $p(\bx)=1$, so the optimal Importance Sampling density that minimizes the variance in estimating $\ell$ is $q^*(\bx)=H(\bx)/\ell$. Therefore, the target function is $h(\bx)=H(\bx)$.

\begin{figure}[H]
    \centering
    \includegraphics[width=\textwidth]{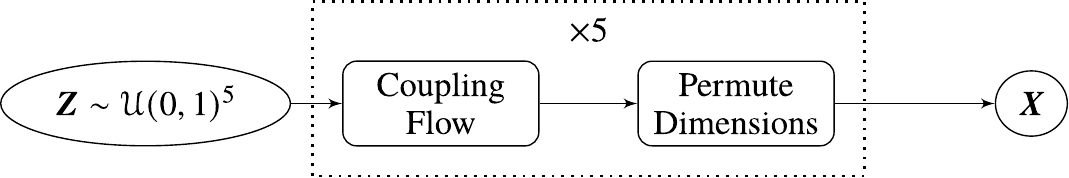}
    \caption{Flow chart showing the normalizing flow structure used in the bridge network example. The base distribution is a 5D uniform distribution which forms the input to a sequence of five consecutive Coupling Flows and dimension permutation pairs.}
    \label{fig:flow_chart_bridge}
\end{figure}

The space of allowed values in this example is $\mathcal{X}=[0,1]^5$, so it is reasonable to choose a model architecture in which the domains and co-domains of each composed flow unit is $[0,1]^5$. To achieve this we choose a uniform base distribution, $Z_i\sim \U(0,1)$. Additionally, the model is composed of five Coupling Flows and dimension permutation pairs. In each permutation the dimensions are cycled via $[3,4,5,1,2]$. The dimensions are partitioned so the coupling function transforms the first three dimensions. The coupling function of each Coupling Flow is a rational function based on \cref{eq:nonlinsquare} with separate parameters for each dimension. The conditioner, like the previous example, is a linear transformation. Figure \ref{fig:flow_chart_bridge} uses a flow chart to illustrate the structure of this model. The parameters in the coupling functions of each Coupling Flow unit contain additional constraints to ensure that $r(0)=0$ and $r(1)=1$, thereby preserving the domain of $[0,1]^5$ at each flow component.

\begin{figure}[H]
    \centering
    \includegraphics[scale = 0.7]{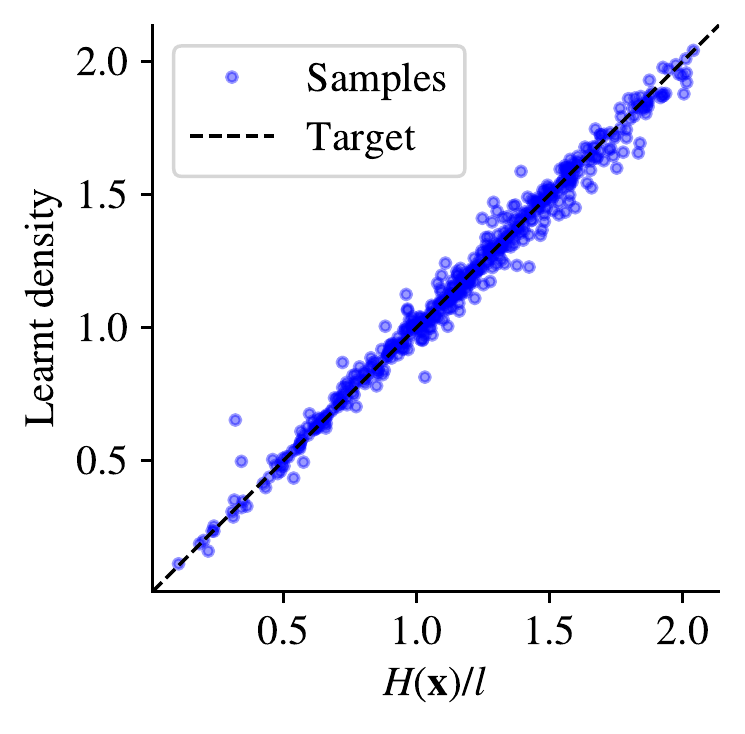}
    \includegraphics[scale = 0.7]{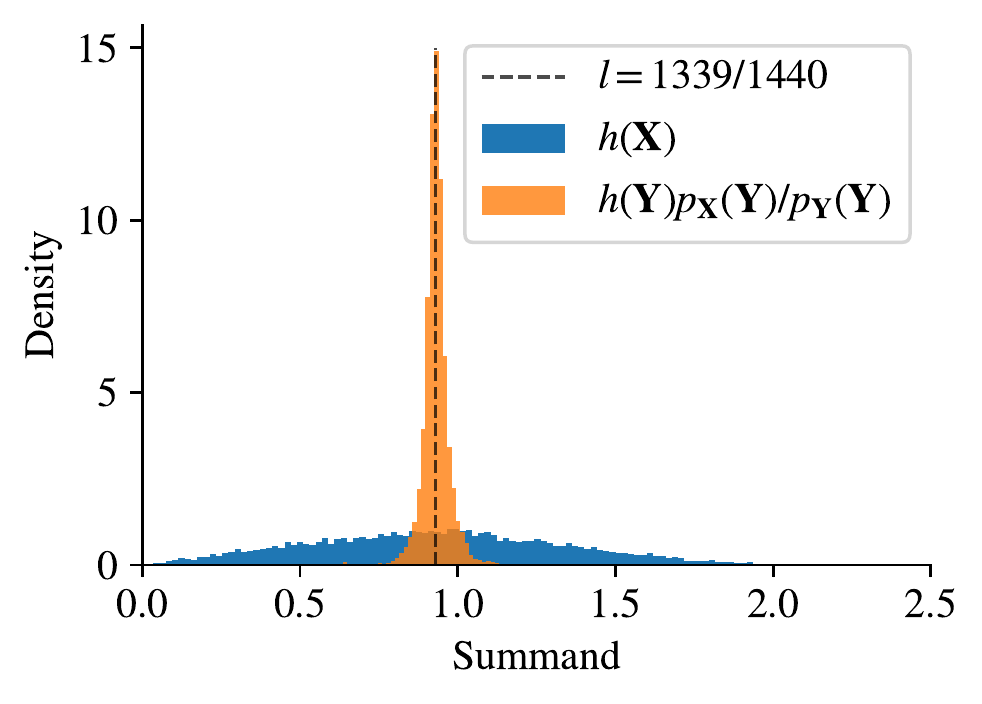}
    \caption{A scatter plot (left) comparing the learned probability density and the target density, and a histogram (right) comparing the summand densities of the crude Monte Carlo estimator (blue) and the Importance Sampling estimator (orange). Clearly the model has learned to approximate the target density and can estimate the expected minimum path length with a much lower variance using Importance Sampling than the crude Monte Carlo estimator.}
    \label{fig:bridge_hist}
\end{figure}

The model was trained with a learning rate of 0.0001, weight decay parameter of 0.0001, batch size 10,000 for 300,000 iterations. The scatter plot in Figure~\ref{fig:bridge_hist} compares the learned probability density of $\v \psi(\bZ;\bu)$ with the optimal sampling  density given by \cref{eq:optimal_dist_H}. The learned density forms a fairly good approximation of the optimal sampling density with a KL divergence of about 0.0017 with 29\% standard error (using sample of 10,000). The histograms in Figure~\ref{fig:bridge_hist} compare the densities of the summands of the Importance Sampling estimator from \cref{eq:expected_lambda} and the crude Monte Carlo estimator, being the sample mean of $H(\bX)$. In this example both methods provide accurate estimates of the expected minimum path length. Typical values of the crude Monte Carlo and IS estimators are 0.9272 with a standard relative error of $0.43\%$ and 0.9301 with a standard relative error of $0.053\%$. While both values agree with the theoretical value of $l = \frac{1339}{1440}=0.92986\dot{1}$, the IS estimator reduces the variance by a factor of about 65.

\subsection{Conditional Bridge Network}
In this example we continue with the bridge network, but now consider
the rare event that the  shortest path includes three edges, passing through the middle, rather than just two. That is, the shortest path is either $ABCD$ or $ACBD$. The rare-event region can be defined by choosing
\[
    S(\bX) = \min\left(X_1+3X_3+2X_5,2X_2+3X_3+X_4\right)-\min\left(X_1+X_4,2X_2+2X_5\right),
\]
and $\gamma = 0$. That is, the penalty depends on the difference between the shortest `inner' path ($ABCD$ or $ACBD$) and the shortest `outer' path ($ABD$ or $ACD$).

The exact same model was reused, but trained across 500,000 iterations with the penalty factor included in the target function. In this way, the goal is to estimate the expected shortest path length of the distribution conditioned on the shortest path being $ABCD$ or $ACBD$. The rare event probability can be estimated using \cref{eq:c_estimator} with a sample size of 10,000 to be about 0.0346 with a standard relative error of $1.3\%$. This corresponds to a variance reduction in a factor of about 17 compared to the crude Monte Carlo estimate. The conditional expected shortest path length can now be estimated via \cref{eq:lc_estimate} as 0.913 with a standard relative error of $1.7\%$.

\begin{figure}[H]
    \centering
    \includegraphics[width = 0.49\textwidth]{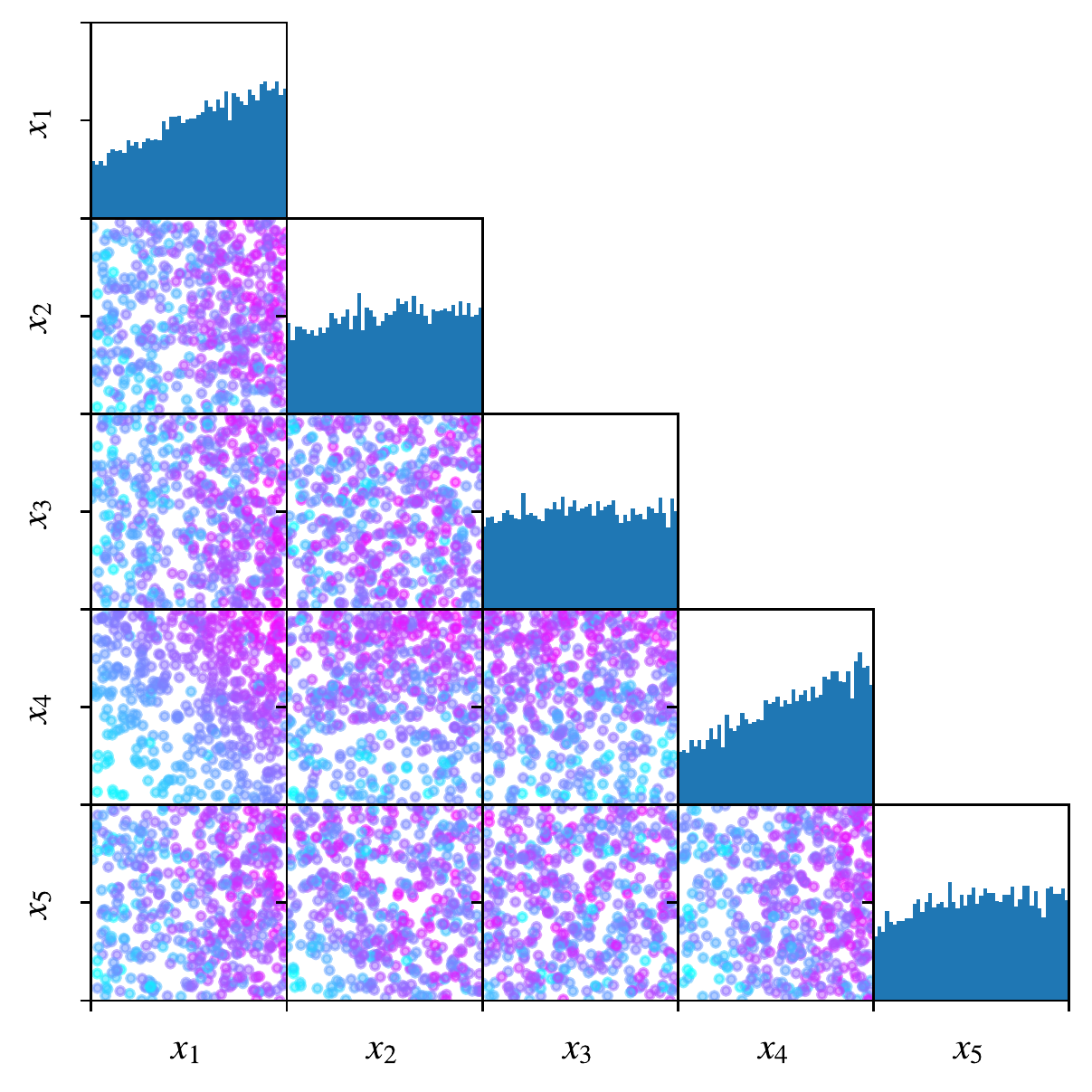}
    \includegraphics[width = 0.49\textwidth]{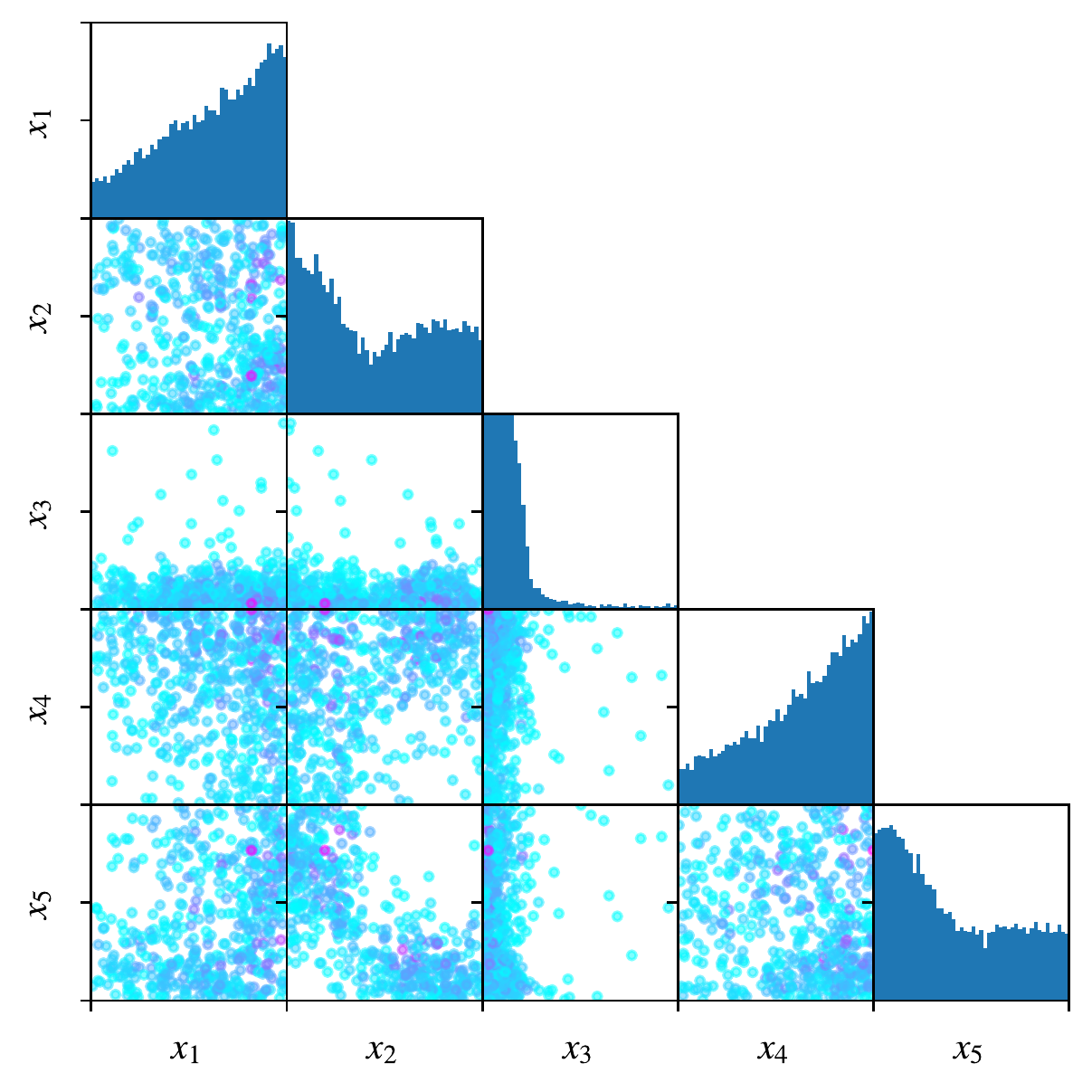}
    \caption{Scatter plots comparing the each of the five dimensions against each other, as well as histograms showing the distribution of each dimension. The darker purple colour corresponds to a higher probability density. The left figure shows the learned unconditional distribution while the right shows the learned conditional distribution.}
    \label{fig:bridge_scatter}
\end{figure}

By learning an approximation of the conditional distribution, we can
explore the likely conditions required to generate the rare
event. Figure~\ref{fig:bridge_scatter} compares scatter plots between
the learned unconditional and the conditional distributions. The unconditional distribution is not too different from the original uniform distribution, with the exception of $X_1$ and $X_4$. This is the expected result since the path, $ABD$, with path length $X_1+X_4$, is the most likely to be the shortest. However, the learned conditional distribution is significantly different to the original uniform distribution. From the scatter plots it is clear that $X_3$ is almost always short, and $X_2$ and $X_5$ have a strong negative correlation. Therefore, we can conclude that the most likely conditions for the shortest path to be one that passes through the middle edge, is that the middle edge is short, and that exactly one of bottom two edges is short. This is congruent with expectations since a long middle edge would likely make the total path length longer than both of the outer paths, if both bottom edges were short then the middle edge would be bypassed, and if both bottom edges were long, then the shortest path would likely be across the top.

\subsection{Asian Call Option}
\begin{figure}[H]
    \centering
    \includegraphics[width=\textwidth]{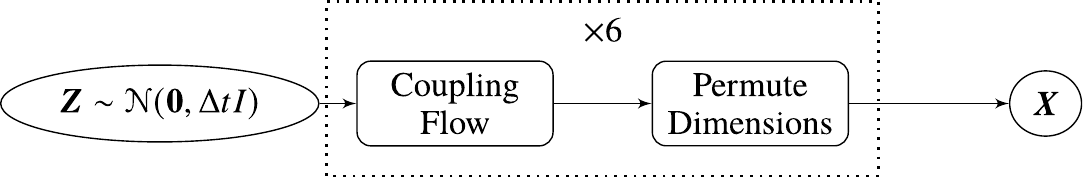}
    \caption{Flow chart showing the normalizing flow structure used in the Asian call option example. The base density is a 88D multivariate normal density which forms the input to a sequence of six consecutive Coupling Flow and dimension permutation pairs.}
    \label{fig:flow_chart_asian_call}
\end{figure}

In this section we look at stock price models and expected pricing of call options. This example, from Chapter~15 of \citet{kroese2013handbook}, specifically looks at estimating the expected Asian call option of a stock whose undiscounted price, $S_t$,  at time $t$ can be modelled by the Stochastic Differential Equation (SDE)
\[
    dS_t=rS_tdt+\sigma S_tdX_t,
\]
where the drift, $r$, is the risk-free rate of return, $\{X_t\}$ is a Wiener process and $\sigma$ is the volatility. This SDE is solved by
\[
    S_t = S_0 \e^{(r-\sigma^2/2)/t+\sigma X_t}.
\]
The average stock price across the interval $t\in[0,T]$ can be approximated by
\[
    \bar{S}_T=\frac{1}{n+1}\sum^n_{i=0}S_{t_i}, \quad t_i = \frac{iT}{n}, ~ i=0,\ldots,n,
\]
where the time has been discretized into $n$ equal intervals of $\Delta t = T/n$,
\[
    S_{t_i}=S_0 \exp\left((r-\sigma^2/2)t_i+\sigma X_{t_i}\right),
\]
and $X_{t_i}\sim \EuScript{N}(0,\Delta t)$. The discounted payoff at time $t=0$ of the Asian option is
\[
    H(\bX)=\e^{-rT}(\bar{S}_T-K)^+,
\]
where $\bX = \left(X_{t_1},\ldots, X_{t_n}\right)$ and the $+$ superscript denotes the ramp function. Since $\{X_t\}$ is a Wiener process, $\bX\sim \Nor(\mathbf{0},\Delta t I)$, where $I$ is the identity matrix of appropriate dimensionality. If the goal is to estimate the expected discounted payoff then, according to \cref{eq:optimal_dist_H}, the optimal sampling distribution should include a factor of $H(\bx)$. However, the objective \cref{eq:objective} requires the target function to be non-zero at all sampled values. Therefore, we introduce the following approximation
\[
    x^+\approx v(x):=
    \begin{cases}
    x, & \text{if }x\geqslant \delta \\
    \delta \e^{\frac{x}{\delta}-1}, & \text{if }x<\delta
    \end{cases}
\]
which is a strictly positive continuously differentiable function that is exact in the limit $\delta\rightarrow 0$. In our experiments we choose $\delta = 0.5$. Therefore, the target function is
\[
    h(\bx)=v\left(\e^{-rT}(\bar{S}_T-K)\right) \frac{1}{(2\pi \Delta t)^{n/2}}\e^{-\frac{1}{2\Delta t}\vert\bx\vert^2}.
\]
We chose the base distribution to be $\bZ\sim \Nor(\mathbf{0},\Delta t I)$ and the normalizing flow to be comprised of six coupling flow units with equal partition sizes. The dimensions were permuted after each coupling flow unit. Figure \ref{fig:flow_chart_asian_call} illustrates the normalizing flows structure.

\begin{figure}[H]
    \centering
    \includegraphics[scale = 0.7]{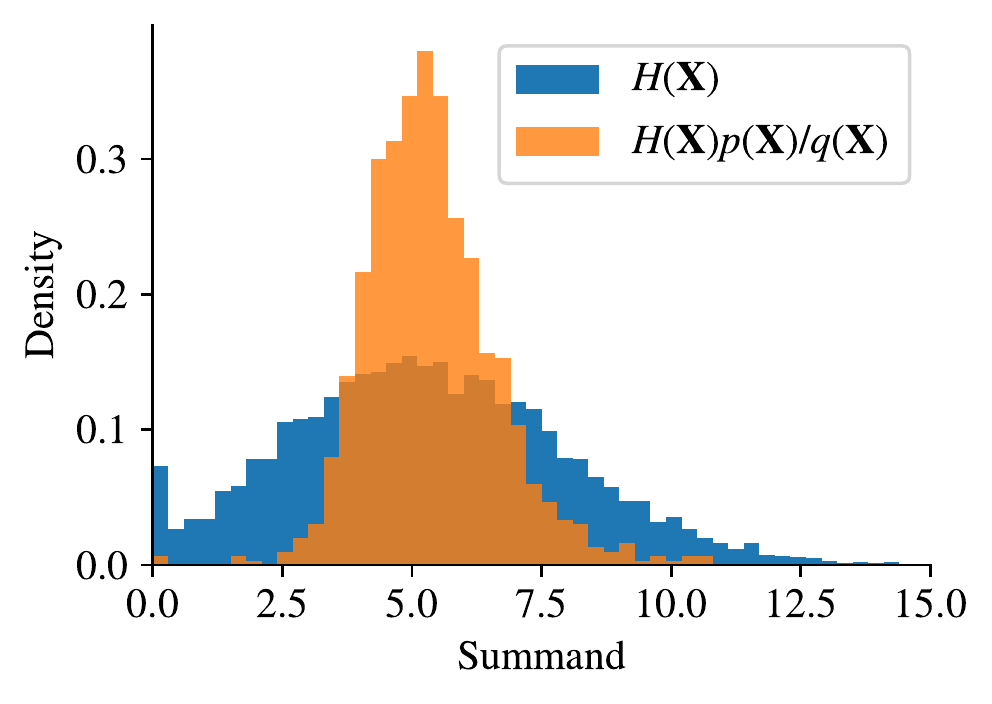}
    \caption{Histograms of the crude Monte Carlo and Importance Sampling estimator summands. Importance Sampling using the learned distribution allows the expected discounted payoff of the Asian option to be estimated with less variance than the crude Monte Carlo estimate.}
    \label{fig:asian_hist}
\end{figure}

The parameters used in this example were $r=0.07$, $\sigma = 0.2$, $K=35$, $S_0=40$, $T=4/12$ and $n=88$. The model was trained for 30000 steps with a batch size of 1000, learning rate of 0.0001 and weight decay of 0.0001. Figure \ref{fig:asian_hist} compares the histograms of the crude Monte Carlo and importance sampling estimator summands. Using a sample of 10000, the crude estimator is about 5.3432 with a standard relative error of $0.49\%$, and the importance sampling estimator is about 5.3580 with a standard relative error of $0.23\%$. This corresponds to a variance reduction of about a factor of 4.4.

Let us explore the rare event in which the discounted payoff exceeds 14, ${\mathcal{A}=\{\bx\in\mathbb{R}^n ~:~ \e^{-rT}(\bar{S}_T-K)^+ \geq \gamma\}}$ with performance function $S(\bx) = \e^{-rT}(\bar{S}_T-K)\geqslant\gamma$ and $\gamma = 14$. In this case the goal is to explore how this rare event occurs and to estimate the probability of the rare event. Therefore, the target function is just the probability density function of the Wiener process multiplied by the penalty factor,
\[
    h(\bx) = \frac{1}{(2\pi\Delta t)^{n/2}}\e^{-\frac{1}{2\Delta t}\vert\bx\vert^2}\rho(\bx).
\]

\begin{figure}
    \centering
    \begin{tabular}{cc}
    \includegraphics[width=0.49\textwidth]{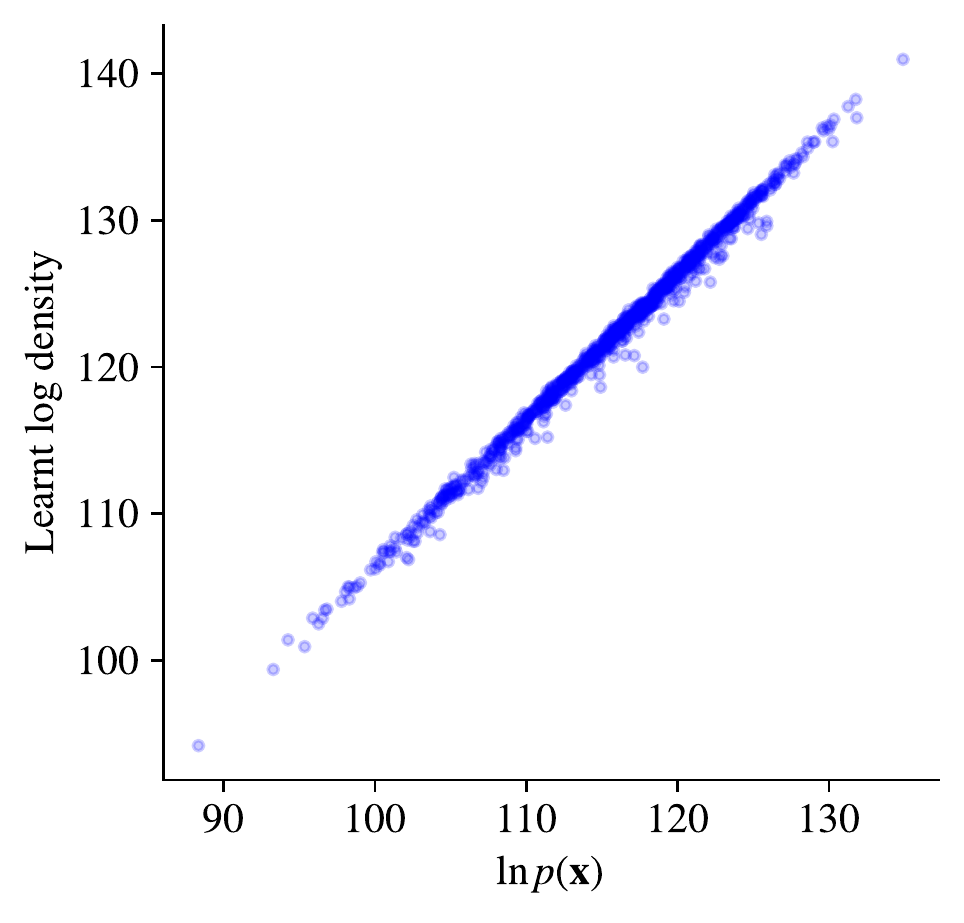} &
    \includegraphics[width=0.49\textwidth]{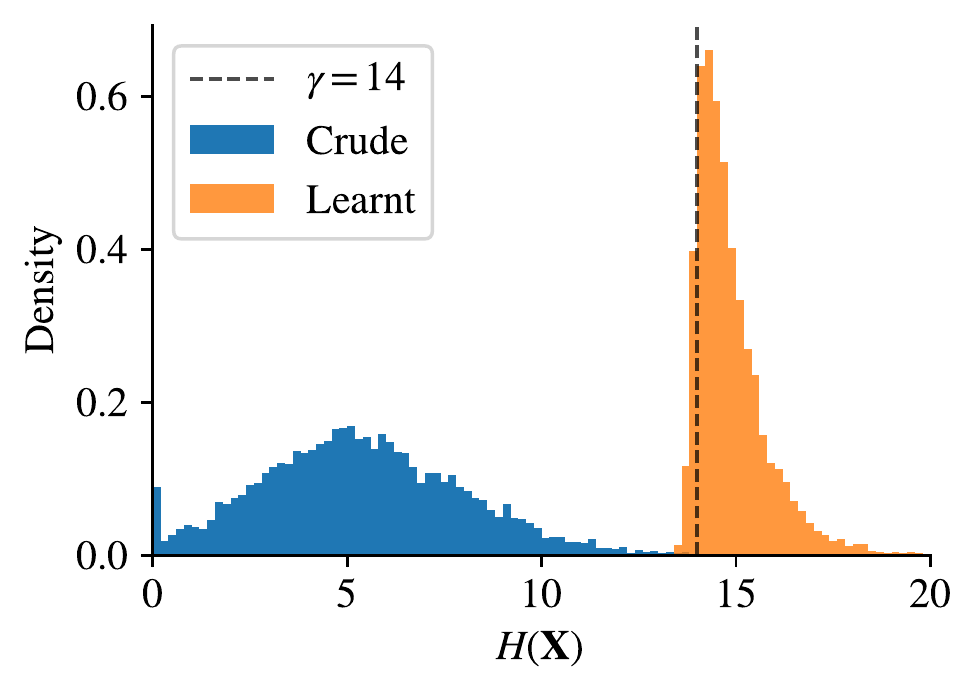}
    \end{tabular}
    \caption{A scatter plot comparing the learnt log probability density and the unconditional log probability density (left) and Histograms (right) of simulated discounted payoffs using crude Monte Carlo (blue) and the learned conditional distribution (orange).}
    \label{fig:asian_learnt_density}
\end{figure}

This time the model was trained for 100000 iterations. With a sample of 10000 the crude and IS estimators are about 0.0022 with $21\%$ relative error, and 0.0015797 with $0.46\%$ relative error respectively. This corresponds to a variance reduction of a factor of about 2100. Using the estimated normalization constant of 0.0015797 the KL divergence between the learnt density and the target density is about 0.15157 with a standard relative error of $0.40\%$ indicating that the learnt distribution is very close to the target conditional distribution. Figure \ref{fig:asian_learnt_density} illustrates the strong linear relationship between the learnt log density and the target unconditional density, and compares the distributions of the simulated discounted Asian options.

\begin{figure}[H]
    \centering
    \begin{tabular}{cc}
    \includegraphics[width = 0.48\textwidth]{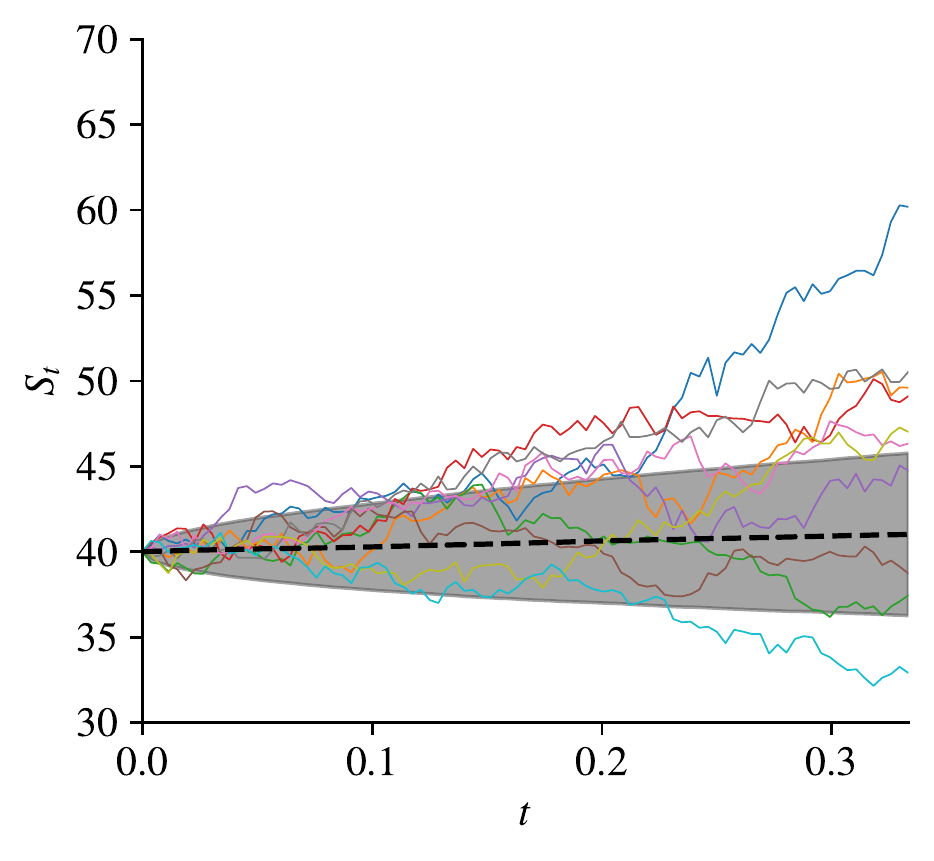} &
    \includegraphics[width = 0.48\textwidth]{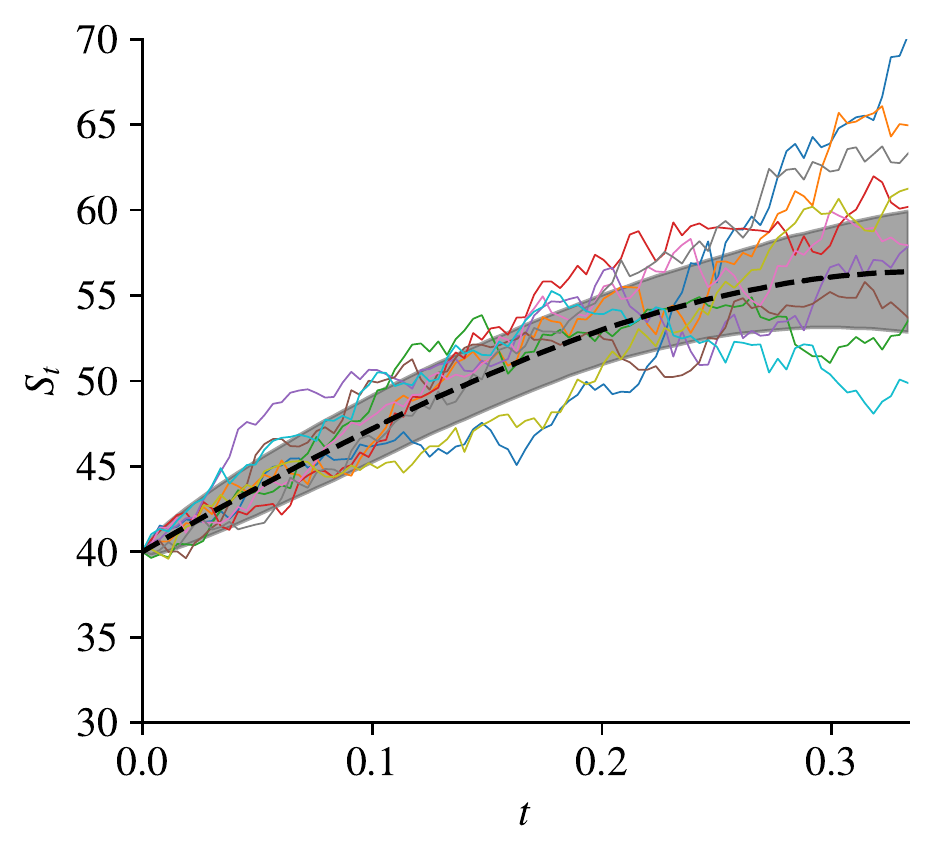}
    \end{tabular}
    \caption{Undiscounted stock prices generated by crude Monte Carlo (left) and the model conditioned on discounted payoffs of at least 14 (right). Coloured lines represent 10 trajectories, dashed black lines represent the mean and the shaded grey region is a single standard deviation from the mean.}
    \label{fig:asian_traj}
\end{figure}

\begin{figure}[H]
    \centering
    \includegraphics[width=\textwidth]{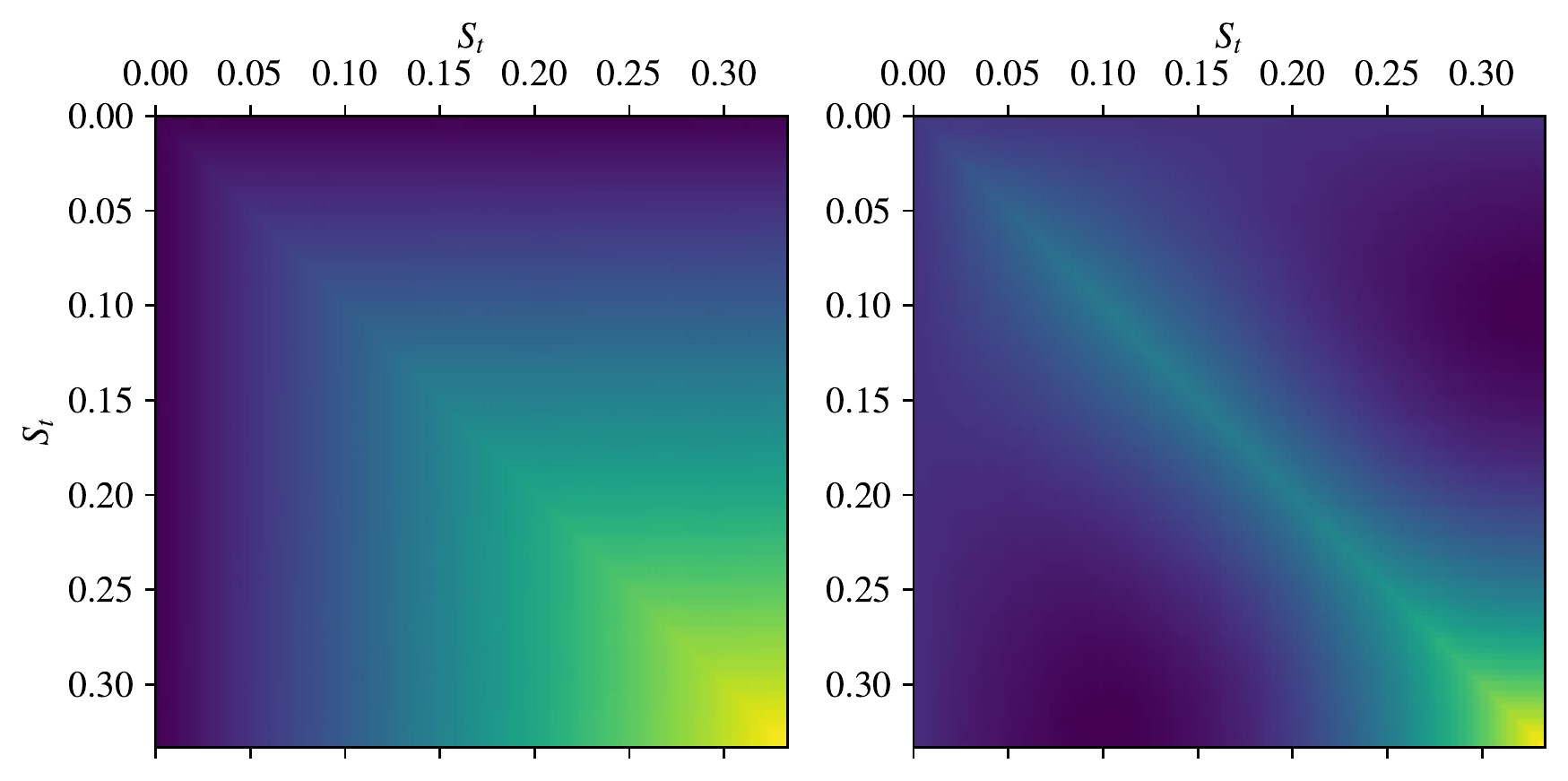}
    \caption{Covariance matrix of undiscounted stock prices simulated using crude Monte Carlo (left) and generated by the model conditioned on discounted payoffs of at least 14 (right). In the crude simulation the variance grows with with time. There is a small amount of covariance between early stock prices and later stock prices in the conditionally generated prices. It appears that while most stock prices achieve the discounted payoff of at least 14 through consistent steady increases, there are a few trajectories which perform well near the beginning or near the end, but not both.}
    \label{fig:asian_cov}
\end{figure}

Since the learnt distribution is a decent approximation of the conditional distribution, we can use the simulated trajectories to gain insights into how high performing options can occur. Figure \ref{fig:asian_traj} shows typical stock prices across time when simulating using the original probability measure and the learnt probability measure. In the first instance the average value is largely static, increasing very slowly, while the variance grows across time. However, conditioning the simulation on achieving a discounted payoff of at least 14 changes the trend unexpectedly. Rather than a steady growth in stock price across the whole time interval, most of the stock price growth occurs by $t=0.25$. Additionally, the variance is fairly constant over time during this growth time, mostly growing near the end. This result is also visible in the covariance matrix of the stock price, illustrated in figure \ref{fig:asian_cov}. Under the original probability distribution the stock price near maturity depends little on the early prices, and the variance grows roughly linearly with time. However, under the learnt probability distribution, the variance in stock price is largely constant until a short time before maturity. Additionally, there is a small correlation between early stock prices and late stock prices. This could indicate that 

\subsection{Double Slit Experiment}
In this example we test our method on a high-dimensional problem. A dimensionless particle traverses an unbounded $2$-dimensional environment for a maximum duration of $T$, which is discretized into $\Delta t = \frac{d}{2}$ time steps, where ${d\geq2}$. The environment consists of a barrier with two slits, located at $(x_{\mathrm{slit}}, y_{\mathrm{slit}})$ and $(x_{\mathrm{slit}}, -y_{\mathrm{slit}})$. The width of both slits is $w_{\mathrm{slit}}$. Additionally, the environment contains a vertical screen located at $x_{\mathrm{screen}}$. The aim of the particle is to reach the screen while simultaneously passing through one of the two slits. At time step $0$, the particle starts at the $2D$-position $\point_0 = (0, 0)$. Given a random vector $\bQ\sim\Nor(\v 0, \frac{2T}{d}I_d)$, where $I_d$ is the $d$-dimensional identity matrix, the position of the particle at time step $1\leq k \leq d/2$ evolves according to
\begin{equation}\label{eq:position_change}
\point_{k} = \point_{k-1} + \bQ_k,
\end{equation}
where $\bQ_k = (Q_{2k-1}, Q_{2k})^{\top}$ is the $k$-th component vector of $\bQ$. The path $\bV$ of the particle induced by $\bQ$ is then the sequence of points $\bV = (\point_k)_{k=1}^{d/2}$, with $\point_k$ defined according to \cref{eq:position_change}.

Let $g_d$ be the density of $\Nor(\v 0, \frac{2T}{d}I_d)$. Our goal is to learn a sampling density that is proportional to $g_d$ truncated to the rare-event region $\mathcal{A} = \left \{\bq\in\mathbb{R}^d: S(\bq) \geq 0 \right \}$, where $\bq$ is an outcome of $\bQ$. The performance function $S(\bq)$ is defined as follows. Suppose is $M$ is the time step where $\bX$ is closest to the screen; that is,
\[
    M :=\argmin_{1\leq k \leq d/2} \left( x_{\mathrm{screen}}-X_k\right),
\]
where $X_k$ is the $x$-coordinate of the particle at time step $k$. Additionally, let $N$ be the first step where $X_{N-1} \leq x_{\mathrm{slit}}\leq X_N$, i.e., where the particle crosses $x_{\mathrm{slit}}$. The performance function of an outcome $\bq$ of $\bQ$, with path $\bv = P(\bq)$, and outcomes $n$ and $m$ of $N$ and $M$ is then defined as
\begin{equation}\label{eq:performance_function}
    S(\bq) = -\bigg(\min\{f_1(y_{n}), f_2(y_{n})\} + f_3(x_{m})\bigg),
\end{equation}
where
\begin{align}\label{eq:penalty_components}
f_1(y_n) &= \left (\left \vert y_n-y_{\mathrm{slit}}\right \vert-\frac{w_{\mathrm{slit}}}{2}\right )\mathbbm{1}_{\left \{\left \vert y_n-y_{\mathrm{slit}} \right \vert> \frac{w_{\mathrm{slit}}}{2}\right \} }, \nonumber \\
f_2(y_n) &= \left (\left \vert y_n + y_{\mathrm{slit}} \right \vert -\frac{w_{\mathrm{slit}}}{2} \right )\mathbbm{1}_{\left \{ \left \vert y_n + y_{\mathrm{slit}} \right \vert > \frac{w_{\mathrm{slit}}}{2} \right \}}, \nonumber \\
f_3(x_m) &= \left (x_{\mathrm{screen}} - x_m \right )\mathbbm{1}_{\left \{ x_{\mathrm{screen}} - x_m > 0 \right \}}.
\end{align}
The functions $f_1$ and $f_2$ in \cref{eq:penalty_components} measure the distances of the $y$-coordinate (i.e., $y_n$) of point $\spoint_{n}$ to the slits and the $\min$ operator in \cref{eq:performance_function} ensures that only the minimum distance to the slits contributes to the performance function. Note that the definitions of $f_1$ and $f_2$ imply that a path successfully crosses a slit if $y_{\mathrm{slit}}-\frac{w_{\mathrm{slit}}}{2}\leq\left\vert y_n \right\vert\leq y_{\mathrm{slit}} + \frac{w_{\mathrm{slit}}}{2}$, even if the path touches the barrier from $\spoint_{n-1}$ to $\spoint_n$. In case the particle does not cross $x_{\mathrm{slit}}$, we set both $f_1$ and $f_2$ to $0$. The function $f_3$ is an additional penalty term which penalizes the $x$-coordinate of the point $\spoint_{m}$ (i.e., $x_m$) to be on the left-hand side of the screen. 

\begin{figure}[H]
    \centering
    \includegraphics[width=\textwidth]{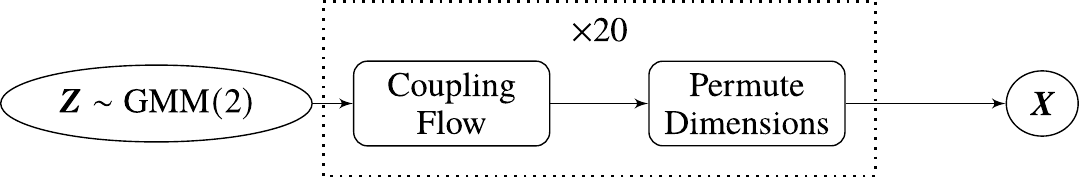}
    \caption{Flow chart showing the normalizing flow structure used in the Double Slit example. The base density is a Gaussian Mixture Model ($\mathrm{GMM}$) consisting of two component distributions $\Nor(\v\mu_1, \frac{2T}{d} I_d)$ and $\Nor(\v\mu_2, \frac{2T}{d} I_d)$ with equal weights, where $\v\mu_1 = (1, -1, \ldots, 1, -1)^{\top}$ and $\v\mu_2 = (1, \ldots, 1)^{\top}$. This forms the input to a sequence of twenty consecutive Coupling Flow and dimension permutation pairs.}
    \label{f:normalizing_flow_double_slit}
\end{figure}

With the performance function in \cref{eq:performance_function}, the target function is defined as
\begin{equation}
h(\bq) = g_d(\bq)\exp\left [-\alpha(\gamma-S(\bq))\mathbbm{1}_{\left \{S(\bq) < 0\right \}}\right ],
\end{equation}
with $\gamma = 0$. For the base distribution, we chose a Gaussian Mixture Model ($\mathrm{GMM}$) consisting of two $d$-dimensional component distributions $\Nor(\v\mu_1, \frac{2T}{d} I_d)$ and $\Nor(\v\mu_2, \frac{2T}{d} I_d)$ with equal weights, where $\v\mu_1 = (1, -1, \ldots, 1, -1)^{\top}$ and $\v\mu_2 = (1, \ldots, 1)^{\top}$. This bimodal mixture model helps in learning the correct bimodal target density. The normalizing flow is composed of twenty coupling flow units with equal partition sizes. The dimensions are permuted after each coupling flow unit. \Cref{f:normalizing_flow_double_slit} illustrates the normalizing flow.

In our experiments we set $T=1$, $x_{\mathrm{slit}} = 5$, $y_{\mathrm{slit}} = 1.5$, $w_{\mathrm{slit}}=1$ and $x_{\mathrm{screen}}=10$. We then trained three normalizing flows using $d=100$, $d=500$ and $d=1,000$, where each flow was trained for $100,000$ iterations with a batch size of $200$, learning rate of $10^{-6}$ and weight decay of $10^{-8}$.

\begin{figure}[H]
    \centering
    \includegraphics[width=0.8\textwidth]{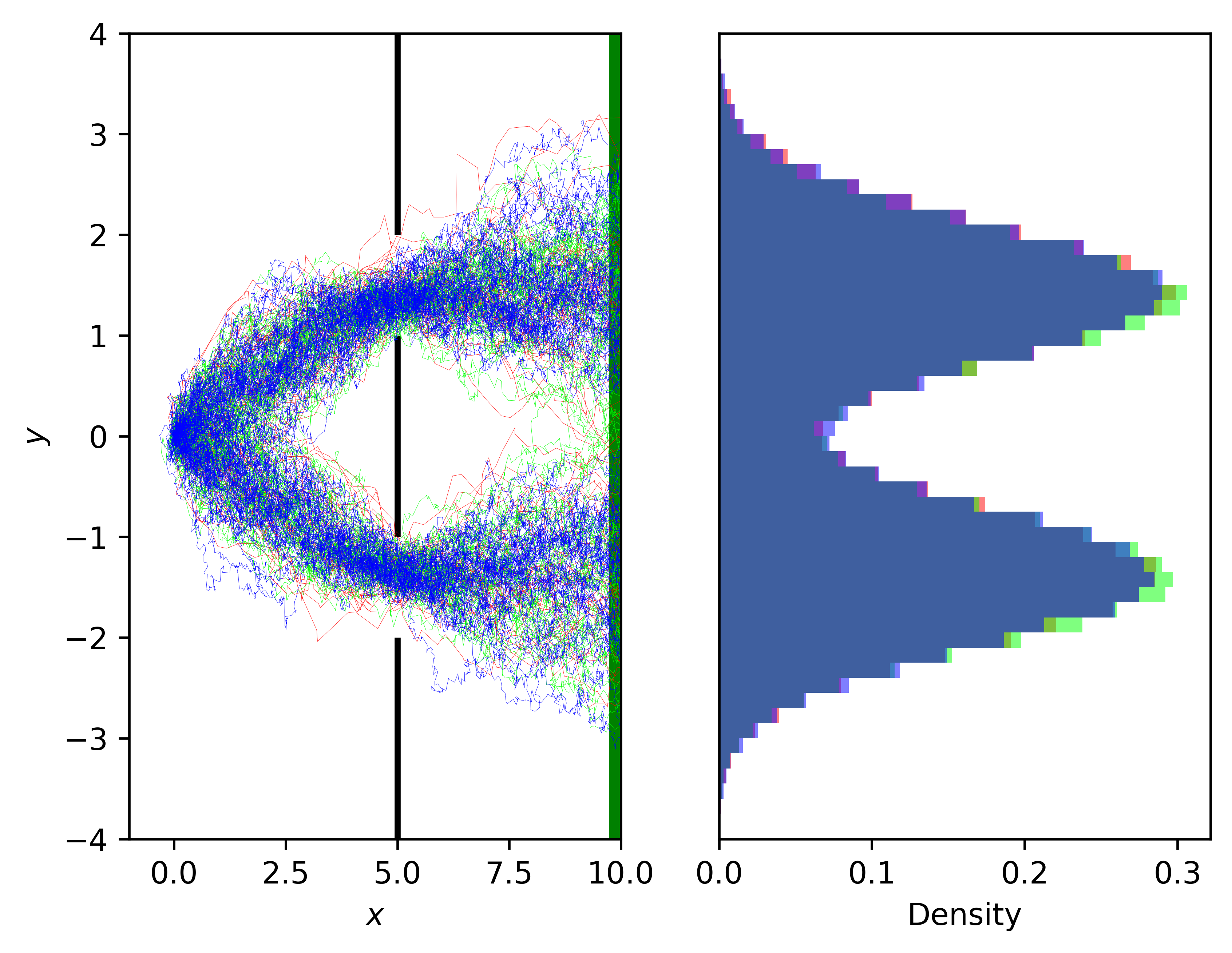}
    \caption{The double slit environment (left figure) with $100$ paths simulated with the learned normalizing flows for $d=100$ (red), $d=500$ (green) and $d=1,000$ (blue). The black regions represent the barrier, while the dark green region represents the screen. The right figure shows the histograms of $100,000$ simulated paths at the screen for $d=100$ (red), $d=500$ (green) and $d=1,000$ (blue).}
    \label{fig:double_slit}
\end{figure}

\Cref{fig:double_slit} (left) shows the double slit environment with paths simulated using the learned normalizing flows for $d=100$ (red paths), $d=500$ (green paths) and $d=1,000$ (blue paths). We additionally used $100,000$ samples for each $d$ to test the capability of the models in generating paths that successfully pass through one of the slits and hit the screen at $x_{\mathrm{screen}}$. In our tests, the success rate was $91.6\%$, $91.2\%$ and $90.3\%$ for $d=100$, $d=500$ and $d=1,000$ respectively, which shows that the learnt distributions are close to the conditional target distribution, even when the base and target distributions are $1,000$-dimensional. We additionally investigate how the marginal distributions of the paths at $x_{\mathrm{screen}}$ look like. \Cref{fig:double_slit} (right) shows the histogram of the $y$-positions of $100,000$ paths at $x_{\mathrm{screen}}$ for $d=100$ (red), $d=500$ (green) and $d=1,000$ (blue) respectively. We can see that for all values of $d$, the distributions at the screen closely match each other. This is both desired and expected, due to the scaling factor $\frac{2T}{d}$ for the variance of the conditional target densities. 

\section{Conclusion}
Monte Carlo methods are important tools for decision making under uncertainty. However, rare events pose significant challenges for standard Monte Carlo methods, due to the low probability of sampling such events. In this paper, we propose a framework for rare-event simulation. Our framework uses a normalizing flow based generative model to learn optimal importance sampling distributions, including conditional distributions. The model is trained using standard gradient descent techniques to minimize the KL divergence between the model density and a function that is proportional to the target density. Empirical evaluations in several domains demonstrate that our framework is able to closely approximate complicated distributions, even in high-dimensional (up to $1,000$-dimensional) rare-event settings. Simultaneously, our framework demonstrates substantial improvements in sample efficiency compared to standard Monte Carlo methods. Given these properties, we believe that combining our framework with Monte Carlo based methods for decision making under uncertainty is a fruitful avenue for future research.


\backmatter

\bmhead{Supplementary information}
The source-code of our framework and the examples is publicly available at \url{https://github.com/hoergems/rare-event-simulation-normalizing-flows}.

\bmhead{Acknowledgements}
We thank Wei Jiang for helpful discussions. This work is supported by the Australian Research Council Centre of Excellence for Mathematical and Statistical Frontiers (ACEMS) CE140100049 grant and the Australian Research Council (ARC) Discovery Project 200101049.

\bibliography{bibliography}


\end{document}